\documentclass{article}

\PassOptionsToPackage{numbers,compress}{natbib}
\usepackage[main,final]{neurips_2026}
\usepackage{enumitem}
\usepackage[utf8]{inputenc} % allow utf-8 input
\usepackage[T1]{fontenc}    % use 8-bit T1 fonts
\usepackage{hyperref}       % hyperlinks
\usepackage{url}            % simple URL typesetting
\usepackage{booktabs}       % professional-quality tables
\usepackage{amsfonts}       % blackboard math symbols
\usepackage{nicefrac}       % compact symbols for 1/2, etc.
\usepackage{microtype}      % microtypography
\usepackage{xcolor}         % colors
\usepackage{graphicx}
\usepackage{svg}
\usepackage{amsmath}
\usepackage{multirow}
\newif\ifsteerrevisions
\steerrevisionsfalse
\newcommand{\rev}[1]{{\ifsteerrevisions\color{red}\fi#1}}
\newenvironment{revision}{\begingroup\ifsteerrevisions\color{red}\fi}{\endgroup}
\title{STEER: Route-Aware Adaptive Reasoning for Autonomous Driving}

\author{%
  \begin{minipage}[t]{\dimexpr\textwidth-2\tabcolsep\relax}
  \centering
  \rev{\bfseries
    \mbox{Yangang Zou$^{1,2,3}$}\hspace{0.7em}
    \mbox{Pei Liu$^{4,5}$}\hspace{0.7em}
    \mbox{Nan Song$^{1,3}$}\hspace{0.7em}
    \mbox{Bozhou Zhang$^{1,3}$}\hspace{0.7em}
    \mbox{Mingyu Guo$^{6,\dagger}$}\hspace{0.7em}
    \mbox{Jun Ma$^{4,5}$}\hspace{0.7em}
    \mbox{Jiankang Deng$^{7}$}\hspace{0.7em}
    \mbox{Xiatian Zhu$^{8,*}$}\hspace{0.7em}
    \mbox{Li Zhang$^{1,2,3,\ddagger}$}\endgraf}
  \vspace{4pt}
  \rev{\normalfont\small
    \mbox{$^{1}$School of Data Science, Fudan University}\quad
    \mbox{$^{2}$Shanghai Innovation Institute}\quad
    \mbox{$^{3}$Logos Intelligence Lab}\quad
    \mbox{$^{4}$HKUST(GZ)}\quad
    \mbox{$^{5}$HKUST}\quad
    \mbox{$^{6}$Li Auto Inc.}\quad
    \mbox{$^{7}$Imperial College London}\quad
    \mbox{$^{8}$University of Surrey}\endgraf}
  \vspace{3pt}
  \rev{\normalfont\small $^{\dagger}$Project leader.\quad $^{*}$Joint last authorship.\quad $^{\ddagger}$Corresponding author.\endgraf}
  \end{minipage}
}

\begin{document}

\maketitle

\begin{abstract}

Vision-language action (VLA) models excel for end-to-end autonomous driving, yet their tendency to over-invoke chain-of-thought (CoT) reasoning mirrors a human cognitive pitfall: \textit{overthinking}. Just as deliberate reasoning can slow or mislead human judgment in routine tasks, excessive CoT in VLA incurs unnecessary compute overhead and can paradoxically degrade planning performance. Recent research tackles this via adaptive reasoning, which dynamically modulates inference complexity by selectively triggering CoT for complex scenes while responding directly to routine ones.
While prior adaptive methods rely on implicit adaptation, recent work shows that explicitly optimizing the route between CoT and direct responses yields superior performance, assuming scene-difficulty annotations which are labor-intensive, poorly scalable, and prone to error due to the subjectivity of judging driving complexity. We explore, for the first time, whether adaptive reasoning with explicit route optimization can be achieved without any scene-difficulty labeling. We introduce \textbf{STEER}, a generic over-reasoning mitigation strategy featuring two key components: (i) \textit{routing uncertainty-aware rollout}, which calibrates sampling to align routing diversity with model uncertainty; and (ii) \textit{cross-route advantage credit assignment}, which introduces a differential metric to reinforce optimal routing decisions based on environmental rewards rather than subjective labels. Extensive experiments on NAVSIM v1/v2 demonstrate that STEER is \rev{competitive with state-of-the-art methods} in both planning quality and the Pareto front of inference efficiency.
\end{abstract}

\section{Introduction}

End-to-end autonomous driving has been significantly advanced by vision-language action (VLA) models, which directly map multimodal observations to driving behaviors~\cite{hu2023planning,jiang2023vad,hwang2024emma,zhou2025autovla}. With large-scale multimodal pretraining, these models introduce stronger semantic grounding and language-based scene understanding into trajectory generation, making it natural to incorporate explicit chain-of-thought (CoT) reasoning into the driving-planning process~\cite{hwang2024emma,yuan2025autodrive,zhang2026minddriverintroducingprogressivemultimodal,zeng2025futuresightdrive}. Such reasoning is particularly valuable in scenarios that require multi-step inference, such as negotiating unprotected intersections, handling dense merges, or reacting to partially occluded hazards.

However, applying CoT reasoning uniformly to every driving scene is far from ideal, a pattern reminiscent of a human cognitive pitfall: \textit{overthinking}. Just as an experienced driver does not consciously deliberate at every familiar junction, forcing explicit reasoning in routine situations can slow down responses, incur unnecessary compute overhead, and paradoxically degrade planning performance~\cite{luo2025adathinkdrive,yang2025r}. These costs are especially concerning in safety-critical autonomous driving, where both decision quality and response efficiency are essential. The key challenge, therefore, is not merely how to reason, but \emph{when} explicit reasoning should be invoked.

% Recent research has begun to deal with this via adaptive reasoning, which dynamically modulates inference complexity. Existing approaches, however, differ substantially in how route selection is handled. AutoVLA~\cite{zhou2025autovla} is the first work of integrating CoT in autonomous driving, but it does not explicitly optimize the route between reasoning and direct response. AdaThinkDrive~\cite{luo2025adathinkdrive} further shows that explicitly optimizing the route between the two response modes yields superior performance, indicating that route selection itself is a critical ingredient for adaptive CoT. However, this explicit routing assumes scene-difficulty annotations. Such supervision is labor-intensive, poorly scalable, and prone to error due to the subjectivity of judging driving complexity; more importantly, perceived scene difficulty is not always aligned with the actual benefit of reasoning. A scene that appears visually crowded may still admit a direct response, while a seemingly simple scene may require nontrivial inference before a safe action can be made.

Recent research has begun to address this challenge through adaptive reasoning, which dynamically modulates inference complexity. Existing approaches, however, differ substantially in how route selection is handled. AutoVLA~\cite{zhou2025autovla} is the first work to integrate CoT into autonomous driving, but it does not explicitly optimize the route between reasoning and direct response. AdaThinkDrive~\cite{luo2025adathinkdrive} further shows that explicitly optimizing this route yields superior performance, suggesting that route selection itself is a critical ingredient for adaptive CoT. However, current explicit-routing approaches typically rely on scene-difficulty annotations to supervise route selection. Such supervision is labor-intensive, poorly scalable, and prone to error due to the subjectivity of judging driving complexity; more importantly, perceived scene difficulty is not always aligned with the actual utility of reasoning. A scene that appears visually crowded may still admit a direct response, while a seemingly simple scene may require nontrivial inference before a safe action can be made.

Motivated by this insight, we explore, for the first time, whether adaptive reasoning with explicit route optimization can be achieved entirely without scene-difficulty labeling. We propose \textbf{STEER}, a generic over-reasoning mitigation approach \rev{without scene-complexity labels}, built within the VLA paradigm.
% To effectively learn route preferences from task rewards, STEER employs a unified training pipeline. First, to prevent the model from prematurely collapsing into the shorter direct response, we introduce \textit{route-balanced supervised fine-tuning} (SFT) to preserve a balanced dual-route prior. Building upon this foundation, 
STEER optimizes the routing policy through two key reinforcement learning (RL) components: (i) \textit{routing uncertainty-aware rollout}, which calibrates sampling to align routing diversity with model uncertainty; and (ii) \textit{cross-route advantage credit assignment}, which introduces a differential metric to reinforce optimal routing decisions based on environmental rewards rather than subjective labels. By contrasting the outcomes across different routes under the same prompt, STEER encourages the model to select the route that yields the highest reward.
To prevent the model from prematurely collapsing into the shorter direct response, we introduce {route-balanced supervised fine-tuning} to preserve a balanced dual-route prior. 

Our contributions are: \textbf{(i)} We identify \textit{overthinking} as a key challenge in VLA-based autonomous driving and, for the first time, explore whether \rev{explicit routing preferences can be learned from comparative task outcomes}, without any scene-difficulty annotations.
\textbf{(ii)} We propose {STEER}, a generic over-reasoning mitigation strategy featuring \textit{route-balanced supervised fine-tuning} to preserve dual-route capabilities, \textit{routing uncertainty-aware rollout} to align sampling diversity with model uncertainty, and \textit{cross-route advantage credit assignment} to reinforce optimal routing decisions directly from environmental rewards, eliminating the reliance on scene-difficulty labels.
\textbf{(iii)} We achieve \rev{results competitive with state-of-the-art methods} on NAVSIM v1/v2, consistently improving planning quality, routing quality, and pushing the Pareto front of inference efficiency over both fixed-routing and previous adaptive reasoning alternatives.

\section{Related work}
\label{sec:related}

\paragraph{Vision-language action models for autonomous driving.}
End-to-end autonomous driving has evolved significantly from modular pipelines~\cite{hu2023planning,jiang2023vad} to unified architectures. Recently, Vision-language Models (VLMs) have been introduced to bridge the gap between semantic reasoning and physical actions~\cite{radford2021learning,bai2025qwen25vltechnicalreport}. Current VLM-based driving approaches broadly fall into two paradigms. The first paradigm utilizes VLMs for scene understanding or high-level reasoning to produce meta-actions that guide downstream planners~\cite{tian2024drivevlm,liao2025diffusiondrive,li2026sgdrivescenetogoalhierarchicalworld}. Although this facilitates semantic integration, it prevents full end-to-end optimization. The second paradigm, VLA models, directly integrates VLMs with action generation to predict final driving trajectories autoregressively~\cite{hwang2024emma,zhou2025autovla,li2026mtdrive}. Our work builds upon this end-to-end VLA paradigm. However, rather than solely focusing on \emph{how} to generate trajectories, we address a critical internal decision process: optimizing \emph{when} the model should invoke explicit reasoning before committing to an action.

\paragraph{Adaptive reasoning in autonomous driving.}
With the rise of long CoT reasoning in large language models, such as DeepSeek-R1~\cite{guo2025deepseek} and OpenAI o1~\cite{jaech2024openai}, explicit reasoning has become an effective tool for complex decision-making. In autonomous driving, methods such as EMMA~\cite{hwang2024emma} show that domain-specific CoT can improve trajectory prediction. However, reasoning is not universally beneficial: in routine driving scenes, forcing CoT can increase latency, introduce unnecessary verbosity, and even degrade planning quality~\cite{luo2025adathinkdrive,yang2025r}. This has motivated adaptive reasoning, where the core challenge is to decide when explicit deliberation is needed. AutoVLA~\cite{zhou2025autovla} addresses this through implicit adaptation: it mainly optimizes trajectory quality with sequence-level reinforcement learning, where supervision is applied to the overall response rather than to route selection itself. AdaThinkDrive~\cite{luo2025adathinkdrive} moves one step further by explicitly supervising routing: it defines scene complexity for route selection and adds a dedicated routing reward to directly optimize the choice between CoT reasoning and direct response. It further shows that explicit route optimization outperforms implicit adaptation. The limitation, however, is that such supervision depends on scene-complexity annotation, which is labor-intensive, difficult to scale, and inherently subjective. Our work addresses this limitation by studying explicit adaptive reasoning \rev{without scene-complexity labels} for autonomous driving.

\section{Method}
\label{sec:method}

% \begin{figure}[!t]
%     \centering
%     \includegraphics[width=1.0\linewidth]{figures/main_0506.png}
%     \caption{\textbf{Overview of STEER.} STEER formulates adaptive reasoning in autonomous driving as an explicit routing problem, where an initial decision selects either a reasoning route or a direct route before trajectory generation. The framework is trained in three stages: (1) driving-centric pretraining for domain adaptation, (2) route-balanced supervised fine-tuning to preserve support for both valid routes, and (3) route-aware policy optimization to learn the routing preference directly from task rewards, without any scene-difficulty labels.}
%     \label{fig:main_figure}
% \end{figure}

\begin{figure}[!t]
    \centering
    \includegraphics[width=1.0\linewidth]{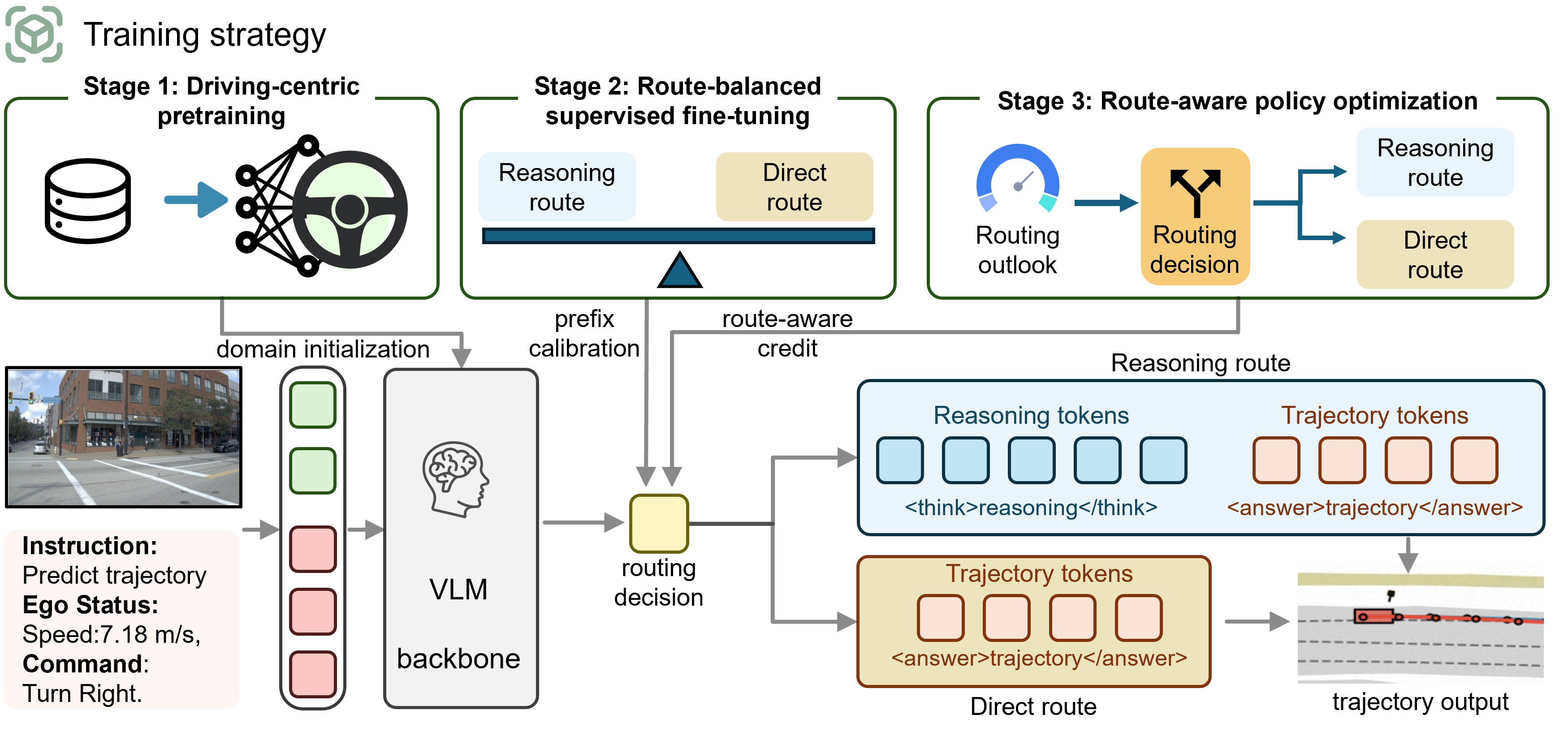}
    \caption{\textbf{Overview of STEER.} STEER is built on an autoregressive vision-language model that takes a front-view driving image together with structured driving context as input, and generates a textual future ego trajectory as output. It formulates adaptive reasoning in autonomous driving as an explicit routing problem, where an initial decision selects either a reasoning route or a direct route before trajectory generation. The framework is trained in three stages: (1) driving-centric pretraining for domain adaptation, (2) route-balanced supervised fine-tuning to preserve support for both valid routes, and (3) route-aware policy optimization to learn the routing preference directly from task rewards, without any scene-difficulty labels.}
    \label{fig:main_figure}
\end{figure}

% \subsection{Problem formulation and model architecture}
% \label{sec:problem_formulation}

{\bf Model formulation.}
To address overthinking in VLA-based autonomous driving, we propose \textbf{STEER}, an adaptive reasoning framework that learns explicit routing between CoT reasoning and direct response from comparative task outcomes without scene-complexity labels.
We instantiate STEER on top of an autoregressive vision-language model for end-to-end driving, e.g., Qwen2.5-VL-8B~\cite{bai2025qwen25vltechnicalreport}. Given a multimodal driving prompt $\mathbf{x}$, the model generates a textual response $\mathbf{y}$ that specifies the future ego trajectory. To explicitly model whether the model should deliberate or respond immediately, we formulate the first step as a discrete routing decision $\mathcal{R} \in \{\texttt{reason}, \texttt{direct}\}$, corresponding to two valid response formats:
\begin{itemize}[leftmargin=*]
    \item {\textit{Reasoning}:} \texttt{<think>} reasoning \texttt{</think>} \texttt{<answer>} trajectory \texttt{</answer>}
    \item {\textit{Direct responding}:} \texttt{<answer>} trajectory \texttt{</answer>}
\end{itemize}
%and $g_{\texttt{direct}}$ denotes entering the direct-response route.

The response distribution is then factorized accordingly as
\begin{equation}
p_\theta(\mathbf{y}\mid\mathbf{x})
=
p_\theta(\mathcal{R}\mid\mathbf{x})
\cdot
p_\theta(\mathbf{y}\mid \mathbf{x}, \mathcal{R}).
\label{eq:factorization}
\end{equation}
The first term determines which route to take, serving as an explicit routing policy, while the second factor generates the driving trajectories conditioned on the selected route. 

% Concretely, each prompt admits two valid response formats:
% \begin{itemize}[leftmargin=*]
%     \item \textbf{Direct route:} \texttt{<answer>} trajectory \texttt{</answer>}
%     \item \textbf{Reasoning route:} \texttt{<think>} reasoning \texttt{</think>} \texttt{<answer>} trajectory \texttt{</answer>}
% \end{itemize}

% This formulation casts route selection $p_\theta(\mathcal{R}\mid\mathbf{x})$ as an explicit policy decision. 

\textbf{Model training.} As illustrated in Figure~\ref{fig:main_figure}, STEER follows a three-stage training pipeline: \textbf{(1)} driving-centric pretraining to adapt the base model to driving tasks  (see Appendix~\ref{app:pretrain} for details), \textbf{(2)} route-balanced supervised fine-tuning (SFT) to preserve the model's prior to execute both responding routes (Section~\ref{sec:sft}), and \textbf{(3)} route-aware policy optimization to optimize the routing policy (Section~\ref{sec:route_aware_po}).

%These stages play complementary roles: pretraining adapts the base model to driving semantics, SFT preserves the model's ability to execute both valid response routes, and RL discovers the actual route preference.

% Our goal is to optimize this routing policy directly using downstream driving rewards, allowing the model to learn which route is more beneficial for a given scene. To achieve this, our training pipeline addresses two key requirements. First, the model must be initialized with a balanced prior that preserves its ability to execute both routes, which we accomplish through route-balanced SFT (Section~\ref{sec:sft}). Second, the RL process must actively explore counterfactual routes and properly assign credit to the routing decision based on comparative outcomes, which we address via route-aware policy optimization (Section~\ref{sec:route_aware_po}).

\subsection{Stage 2: Route-balanced supervised fine-tuning}
\label{sec:sft}

Before RL can discover \emph{when} explicit reasoning is beneficial, the model must first retain the capacity to execute \emph{both} valid response routes. However, a vanilla SFT~\cite{ouyang2022training} objective is insufficient for this purpose. Because the direct route is substantially shorter than the reasoning route, standard teacher forcing introduces a strong optimization bias toward the shorter sequence. This length imbalance causes the model to prematurely collapse into the direct response, suppressing its reasoning ability before policy optimization even begins.

To prevent this early collapse, we introduce \emph{route-balanced supervised fine-tuning}. For each prompt $\mathbf{x}$, we provide both a reasoning-route target $\mathbf{y}^{\texttt{reason}}$ and a direct-route target $\mathbf{y}^{\texttt{direct}}$, and optimize the model using a unified loss function:
\begin{equation}
\mathcal{L}_{\texttt{sft}}(\mathbf{x}) =
\underbrace{-\frac{\lambda}{2} \sum_{\mathcal{R}} \log p_\theta(g_\mathcal{R} \mid \mathbf{x})}_{\text{Routing Calibration}}
+ \sum_{\mathcal{R}} \frac{\alpha_\mathcal{R}}{|\mathbf{y}^{\mathcal{R}}|} \sum_{t=1}^{|\mathbf{y}^{\mathcal{R}}|} -\log p_\theta(y_t^{\mathcal{R}} \mid \mathbf{x}, \rev{g_{\mathcal{R}},} \mathbf{y}_{<t}^{\mathcal{R}}).
\label{eq:sft_loss}
\end{equation}
Here, $g_{\mathcal{R}}$ denotes the route-indicating initial token corresponding to route $\mathcal{R}$, and $\alpha_\mathcal{R}$ balances the two routes ($\alpha_{\texttt{reason}} = \alpha$, $\alpha_{\texttt{direct}} = 1-\alpha$). \rev{The sequence $\mathbf{y}^{\mathcal{R}}$ denotes the continuation after $g_{\mathcal{R}}$, which remains in the autoregressive prefix for every continuation token.}

This objective addresses the pitfalls of vanilla SFT through two key designs: \textit{(1) Routing calibration:} The first term explicitly regularizes the initial decision step toward a uniform distribution, ensuring the model maintains non-trivial support for both routes. \textit{(2) Length normalization:} The second term divides the next-token loss by the route-specific length $|\mathbf{y}^\mathcal{R}|$. This prevents the optimization gradients from being skewed by the massive length disparity, ensuring the shorter direct route is not treated as an optimization shortcut. Ultimately, this balanced initialization serves a preparatory role. It preserves the dual-route capability, leaving the actual decision of \emph{when} to reason entirely to the subsequent reward-driven RL stage. 

\subsection{Stage 3: Route-aware policy optimization}
\label{sec:route_aware_po}

\begin{figure}[!t]
    \centering
    \includegraphics[width=1.0\linewidth]{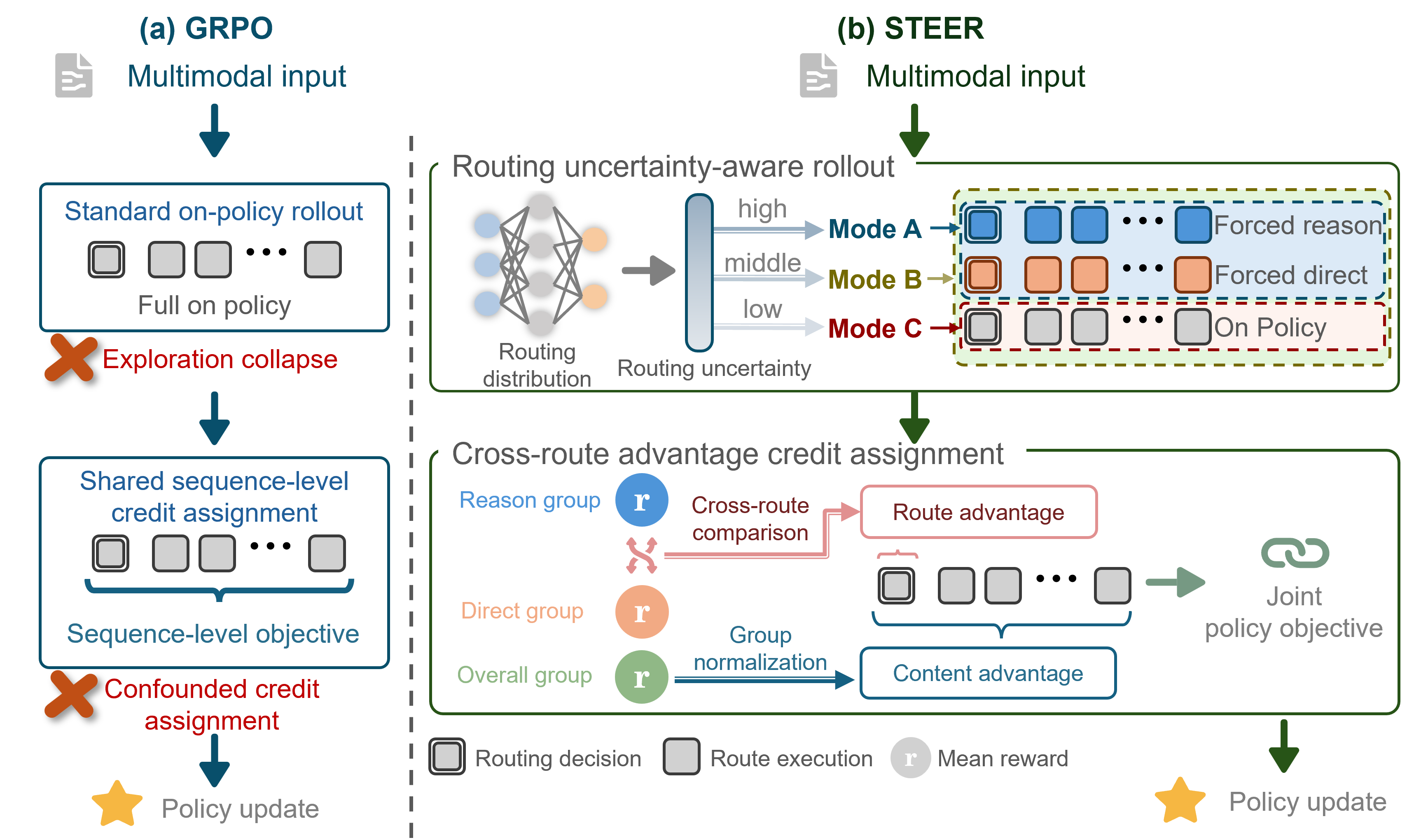}
    \caption{\textbf{Route-aware policy optimization in STEER.} For each prompt, STEER first inspects the initial routing distribution to estimate routing uncertainty. When the routing decision is uncertain, the model performs route-contrastive rollout by explicitly sampling both the reasoning and direct routes under the same prompt; otherwise, it falls back to standard on-policy rollout. \rev{During optimization, STEER applies a standard content advantage to all response tokens and adds a cross-route advantage at the initial routing position, jointly optimizing response quality and route preference.}}
    \label{fig:egrpo_overview}
\end{figure}

Starting from the balanced SFT model, we optimize the routing policy using sequence-level reinforcement learning (e.g., GRPO~\cite{shao2024deepseekmath}) by introducing two key designs: \emph{routing uncertainty-aware rollout} and \emph{cross-route advantage credit assignment} (Figure~\ref{fig:egrpo_overview}).

\paragraph{Routing uncertainty-aware rollout.}
In vanilla GRPO, a group of responses is generated for each prompt via standard on-policy sampling. However, learning \emph{when} to reason requires counterfactual comparison—observing outcomes from both the reasoning and direct routes under the same prompt. Standard on-policy sampling quickly collapses into the slightly preferred route, halting the exploration needed to discover true route utility. 

Given prompt $\mathbf{x}$, the first generated token serves as the routing token. Specifically, if this token is the reasoning special token (i.e., ``\texttt{<t}''), the model takes the reasoning route; if it is the direct special token (i.e., ``\texttt{<}''), the model takes the direct route. We compute the initial routing probabilities $p_{\texttt{reason}}$ and $p_{\texttt{direct}}$ \rev{by normalizing the model probabilities over these two valid route tokens}, and quantify the routing uncertainty using binary entropy:
\[
\mathcal{H}(\mathbf{x}) = -p_{\texttt{reason}}\log p_{\texttt{reason}} - p_{\texttt{direct}}\log p_{\texttt{direct}},
\]
and define the minority-route probability as $p_{\min}(\mathbf{x}) = \min(p_{\texttt{reason}}, p_{\texttt{direct}})$. For a budget of $n$ samples, we dynamically allocate the rollout regime:
\begin{equation}
\label{eq:mode_allocation}
(\mathrm{Mode}, n_{\texttt{reason}}, n_{\texttt{direct}}, n_{\texttt{on}})
=
\begin{cases}
(\mathbf{A}, \lfloor n/2 \rfloor, \lfloor n/2 \rfloor, 0), &
p_{\min} \ge \epsilon,\ \mathcal{H}(\mathbf{x}) > \tau_{\text{high}},\\
(\mathbf{B}, \lfloor n/4 \rfloor, \lfloor n/4 \rfloor, \lfloor n/2 \rfloor), &
p_{\min} \ge \epsilon,\ \tau_{\text{low}} \le \mathcal{H}(\mathbf{x}) \le \tau_{\text{high}},\\
(\mathbf{C}, 0, 0, n), & \text{otherwise},
\end{cases}
\end{equation}
where $n_{\texttt{reason}}$, $n_{\texttt{direct}}$, and $n_{\texttt{on}}$ denote the numbers of forced-reasoning, forced-direct, and standard on-policy rollouts. 

Modes A and B force route-contrastive exploration (manually setting the initial token $g$ and sampling autoregressively) when the routing decision is uncertain ($\mathcal{H}(\mathbf{x})$ is high). Mode C reverts to standard on-policy rollout when the policy is confident, avoiding wasteful exploration.

\paragraph{Cross-route advantage credit assignment.}
Vanilla GRPO evaluates the generated group by computing a shared sequence-level advantage for all tokens in a response based on its relative reward. Without scene-difficulty labels, the model must infer route preference entirely from these task rewards. However, applying a shared advantage confounds the initial discrete routing choice with the much longer trajectory execution, making it impossible to distinguish whether a low reward stems from an unnecessary route or a poorly executed trajectory. 

To resolve this confounding without relying on scene-difficulty labels, we derive the routing preference directly from comparative environmental rewards. \rev{By adding a dedicated cross-route learning signal to the initial decision, we explicitly optimize route preference alongside response quality.}

For prompt $\mathbf{x}_k$, let $\mathcal{G}_k^{\texttt{reason}}$ and $\mathcal{G}_k^{\texttt{direct}}$ be the subsets of reasoning and direct rollouts, with route-wise mean rewards $\bar{r}_k^{\texttt{reason}}$ and $\bar{r}_k^{\texttt{direct}}$, and group mean $\bar{r}_k$. For the initial routing decision $g$, we define a \emph{cross-route advantage} that explicitly compares the realized route against its counterfactual alternative:
\begin{equation}
A_i^{\texttt{route}} =
\begin{cases}
r_i - \bar{r}_k^{\texttt{direct}}, &
i \in \mathcal{G}_k^{\texttt{reason}},\ |\mathcal{G}_k^{\texttt{direct}}| > 0,\\
r_i - \bar{r}_k^{\texttt{reason}}, &
i \in \mathcal{G}_k^{\texttt{direct}},\ |\mathcal{G}_k^{\texttt{reason}}| > 0,\\
r_i - \bar{r}_k, & \text{otherwise}.
\end{cases}
\label{eq:route_advantage}
\end{equation}
\rev{We analyze the expected routing update induced by Eq.~\ref{eq:route_advantage} to characterize its effect on route preference. For a fixed prompt, let $\mu_{\mathrm r},\mu_{\mathrm d}$ be the expected rewards of reasoning and direct responses, $\Delta=\mu_{\mathrm r}-\mu_{\mathrm d}$ their difference, and $p=\pi_\theta(\mathcal R=\texttt{reason}\mid\mathbf x)=\sigma(z)$ the reasoning probability. The sampled route proportions $q_{\mathrm r},q_{\mathrm d}$ can differ from the model probabilities because rollout allocation can force either route.}

\rev{With both routes represented, a simplified population analysis gives:}
\begin{equation}
\rev{\mathbb{E}\!\left[\partial_z\log\pi_\theta(\mathcal R\mid\mathbf x)A^{\texttt{route}}\right]=\left[q_{\mathrm r}(1-p)+q_{\mathrm d}p\right]\Delta.}
\label{eq:route_expected_update}
\end{equation}
\rev{For $0<p<1$, the coefficient multiplying $\Delta$ is positive, so this additional update increases the probability of the route with higher expected reward. The derivation and assumptions are provided in Appendix~\ref{app:route_analysis}.}

\paragraph{Optimization objective.}
\rev{We combine this route-specific signal with standard response optimization. All response tokens, including the routing token, receive the group-relative normalized advantage $A_i^{\texttt{content}} = (r_i-\bar{r}_k)/(\sigma_k+\delta)$, while $A_i^{\texttt{route}}$ provides an additional update at the routing position.}

\rev{Let $\rho_i^{(t)} = p_\theta(y_{i,t}\mid\mathbf{x}_i,\mathbf{y}_{i,<t})/p_{\theta_{\mathrm{old}}}(y_{i,t}\mid\mathbf{x}_i,\mathbf{y}_{i,<t})$ be the token probability ratio, where $y_{i,0}=g_i$ and $\theta_{\mathrm{old}}$ denotes the frozen pre-update parameters. We optimize all response tokens using a standard clipped surrogate objective:}
\begin{equation}
\mathcal{L}^{\texttt{content}}
=
-\frac{1}{|\mathcal{B}|}\sum_{i\in\mathcal{B}}
\sum_{t=0}^{T_i}
\frac{m_i^{\texttt{resp}}(t)}{Z_i}
\min\!\Big(
\rho_i^{(t)} A_i^{\texttt{content}},
\mathrm{clip}(\rho_i^{(t)},1-\epsilon_c,1+\epsilon_c)A_i^{\texttt{content}}
\Big),
\label{eq:content_obj}
\end{equation}
where \rev{$m_i^{\texttt{resp}}(t)$ equals one for response tokens, including the initial routing token ($m_i^{\texttt{resp}}(0)=1$), and zero otherwise}; $Z_i = \sum_t m_i^{\texttt{resp}}(t)$ \rev{counts the response tokens}.

\rev{For the routing decision, we additionally apply a clipped surrogate at the first response position using $A_i^{\texttt{route}}$:}
\begin{equation}
\mathcal{L}^{\texttt{route}}
=
-\frac{1}{|\mathcal{B}|}\sum_{i\in\mathcal{B}}
\min\!\Big(
\rho_i^{(0)} A_i^{\texttt{route}},
\mathrm{clip}(\rho_i^{(0)},1-\epsilon_c,1+\epsilon_c)A_i^{\texttt{route}}
\Big).
\label{eq:route_obj}
\end{equation}
This objective is the core driver of our \rev{adaptation without scene-complexity labels}. By updating the initial routing probability based on the comparative advantage, it directly optimizes the route selection policy $p_\theta(\mathcal{R}\mid\mathbf{x})$.

Combined with a KL regularization term $\mathcal{L}^{\mathrm{KL}}$, the final joint objective is
\begin{equation}
\mathcal{L}
=
\mathcal{L}^{\texttt{content}}
+ \lambda_r \mathcal{L}^{\texttt{route}}
+ \beta_{\mathrm{KL}} \mathcal{L}^{\mathrm{KL}}.
\label{eq:joint_obj}
\end{equation}

\section{Experiments}
\label{experiment}

{\bf Datasets and evaluation metrics.}
We evaluate STEER on NAVSIM v1 and NAVSIM v2~\cite{dauner2024navsim} using the official simulator. Both route-balanced SFT and RL are trained on data constructed from the official navtrain split. For driving-centric pretraining, we additionally use public driving-reasoning datasets, including ReCogDrive~\cite{li2025recogdrive}, LingoQA~\cite{marcu2024lingoqa}, NuScenes-QA~\cite{qian2024nuscenes}, DriveLM~\cite{sima2024drivelm}, and OmniDrive~\cite{wang2025omnidrive}.

Following the official protocol, we use the \rev{Predictive Driver Model Score} (PDMS) on NAVSIM v1 and \rev{Extended Predictive Driver Model Score} (EPDMS) on NAVSIM v2 as the primary planning metrics. These composite scores evaluate trajectory quality by aggregating multiple sub-metrics, such as No Collision (NC), Drivable Area Compliance (DAC), and Ego Progress (EP). To evaluate routing quality \rev{without scene-complexity labels}, we report RouteAcc, AUC, and AvgRegret. These metrics are computed against the \emph{oracle route} (i.e., the route that yields the higher planning score for a given prompt), measuring route-selection correctness, route-ranking quality, and the performance gap to the oracle, respectively. Additionally, to analyze the route balance preserved during the SFT stage, we report the average routing probabilities and routing uncertainty. Full metric definitions for both driving and routing are provided in Appendix~\ref{app:metric}.

{\bf Implementation details.}
STEER is initialized from Qwen2.5-VL-8B~\cite{bai2025qwen25vltechnicalreport} and directly auto-regresses text waypoint trajectories without an additional planner. All training stages use full-parameter fine-tuning and are implemented with DeepSpeed ZeRO-3~\cite{rajbhandari2020zero} on 64 NVIDIA H20 GPUs. Detailed training hyperparameters are provided in Appendix~\ref{app:impl}. \rev{For repeated evaluations, the main tables report means; standard deviations and evaluation details are provided in Appendix~\ref{app:eval_statistics}.}

\subsection{Main results}

%--------------------------------------------------------------
\begin{table*} [!t]
\centering
\caption{\textbf{Performance on NAVSIM v1} navtest using closed-loop metrics. %STEER achieves the best PDMS among autoregressive VLM-based driving methods. 
$^*$ denotes fine-tuned on the NAVSIM trajectory dataset. $\dagger$ denotes Best-of-$N$ evaluation with six generated candidates. }
\resizebox{\textwidth}{!}{
\begin{tabular}{l c c c c c c c c}
\toprule
& Sensors & Trajectory & NC$\uparrow$ & DAC$\uparrow$ & EP$\uparrow$ & TTC$\uparrow$ & C$\uparrow$ & PDMS$\uparrow$ \\
\midrule
\textit{\textbf{Human GT}} & & & & & & & & \\
\textcolor{gray}{Human~\cite{dauner2024navsim}} & - & - & \textcolor{gray}{100.0} & \textcolor{gray}{100.0} & \textcolor{gray}{87.5} & \textcolor{gray}{100.0} & \textcolor{gray}{99.9} & \textcolor{gray}{94.8} \\
\midrule
\textit{\textbf{End-to-End Methods}} & & & & & & & & \\
Ego-MLP~\cite{li2024ego} & 6x C & Continuous & 93.1 & 78.3 & 63.2 & 84.0 & \textbf{100.0} & 66.4 \\
UniAD~\cite{hu2023planning} & 6x C & Continuous & 97.8 & 91.9 & 78.8 & 92.9 & \textbf{100.0} & 83.4 \\
PARA-Drive~\cite{weng2024drive} & 6x C & Continuous & 97.9 & 92.4 & 79.3 & 93.0 & 99.8 & 84.0 \\
Hydra-MDP~\cite{li2024hydra} & 3x C + L & Continuous & 98.3 & 96.0 & 78.7 & 94.6 & \textbf{100.0} & 86.5 \\
DiffusionDrive~\cite{liao2025diffusiondrive} & 3x C + L & Continuous & 98.2 & 96.2 & 82.2 & 94.7 & \textbf{100.0} & 88.1 \\
DiffusionDriveV2~\cite{zou2025diffusiondrivev2} & 3x C + L & Continuous & 98.3 & \textbf{97.9} & 87.5 & 94.8 & 99.9 & 91.2 \\
\midrule
\textit{\textbf{VLA (Planner-assisted VLM Methods)}} & & & & & & & & \\
ImagiDrive-S~\cite{li2025imagidrive} & 1x C & Continuous & \textbf{98.6} & 96.2 & 80.5 & 94.5 & \textbf{100.0} & 87.4 \\
ReCogDrive~\cite{li2025recogdrive} & 3x C & Continuous & 97.9 & 97.3 & 87.3 & 94.9 & \textbf{100.0} & 90.8 \\
SGDrive~\cite{li2026sgdrivescenetogoalhierarchicalworld} & 1x C & Continuous & \textbf{98.6} & 97.8 & 85.8 & 96.2 & \textbf{100.0} & 91.1 \\
\midrule
\textit{\textbf{VLA (Autoregressive VLM Methods)}} & & & & & & & & \\
Qwen2.5-VL$^*$~\cite{bai2025qwen25vltechnicalreport} & 1x C & Text Waypoint & 97.8 & 92.1 & 78.3 & 92.8 & \textbf{100.0} & 83.3 \\
InternVL3$^*$~\cite{zhu2025internvl3} & 1x C & Text Waypoint & 97.0 & 92.4 & 78.9 & 91.8 & \textbf{100.0} & 83.3 \\
AutoVLA~\cite{zhou2025autovla} & 3x C & Discrete Action & 98.4 & 95.6 & 81.9 & \textbf{98.0} & 99.9 & 89.1 \\
AdaThinkDrive~\cite{luo2025adathinkdrive} & 1x C & Text Waypoint & 98.4 & 97.8 & 84.4 & 95.2 & \textbf{100.0} & 90.3 \\
Curious-VLA~\cite{chen2026devil} & 1x C & Text Waypoint & 98.4 & 96.9 & 88.5 & 97.9 & 98.1 & 90.3 \\
\midrule
STEER & 1x C & Text Waypoint & 97.4 & \rev{97.3} & \rev{90.2} & 92.3 & \textbf{100.0} & 91.2 \\
STEER$^\dagger$ & 1x C & Text Waypoint & 98.0 & \rev{97.7} & \rev{\textbf{91.6}} & \rev{94.4} & \textbf{100.0} & \textbf{93.4} \\
\bottomrule
\end{tabular}
}
\label{tab:navsim_v1_main}
\end{table*}
%--------------------------------------------------------------

%--------------------------------------------------------------
\begin{table*} [!t]
\centering
\caption{\textbf{Performance on NAVSIM v2} navtest with extended metrics.}
\label{tab:navsim_v2}
\resizebox{\textwidth}{!}{
\begin{tabular}{l | c c c c c c c c c | c}
\toprule
Method & NC $\uparrow$ & DAC $\uparrow$ & DDC $\uparrow$ & TLC $\uparrow$ & EP $\uparrow$ & TTC $\uparrow$ & LK $\uparrow$ & HC $\uparrow$ & EC $\uparrow$ & EPDMS $\uparrow$ \\
\midrule
VADv2~\cite{chen2024vadv2} & 97.3 & 91.7 & 98.2 & 99.9 & 77.6 & 92.7 & 66.0 & \textbf{100} & 97.4 & 76.6 \\
TransFuser~\cite{prakash2021multi} & 97.7 & 92.8 & 98.3 & 99.9 & 79.2 & 92.8 & 67.6 & \textbf{100} & 95.3 & 77.8 \\
HydraMDP~\cite{li2024hydra} & 97.5 & 96.3 & 98.3 & 99.9 & 80.1 & 93.0 & 65.5 & \textbf{100} & 97.4 & 79.8 \\
HydraMDP++~\cite{li2024hydra} & 97.9 & 96.5 & 98.9 & \textbf{100} & 79.2 & 93.4 & 67.2 & \textbf{100} & 97.7 & 80.6 \\
ARTEMIS~\cite{feng2025artemis} & 98.3 & 95.1 & 98.6 & 99.8 & 81.5 & 97.4 & 96.5 & \textbf{100} & \textbf{98.3} & 83.1 \\
ReCogDrive~\cite{li2025recogdrive} & 98.3 & 95.2 & \textbf{99.5} & 99.8 & 87.1 & 97.5 & 96.6 & 98.3 & 86.5 & 83.6 \\
DiffusionDrive~\cite{liao2025diffusiondrive} & 98.2 & 95.9 & 99.4 & 99.8 & 87.5 & 97.3 & 96.8 & 98.3 & 87.7 & 84.5 \\
Curious-VLA~\cite{chen2026devil} & 98.4 & 96.9 & 99.2 & 99.8 & 88.5 & \textbf{97.9} & \textbf{96.9} & 98.1 & 81.5 & 85.3 \\
SGDrive~\cite{li2026sgdrivescenetogoalhierarchicalworld} & \textbf{98.6} & 94.3 & \textbf{99.5} & 99.9 & 86.0 & \textbf{97.9} & 96.1 & 98.3 & 85.9 & 86.2 \\
\midrule
STEER & 97.4 & \textbf{97.6} & \rev{98.1} & \rev{99.2} & \textbf{95.6} & \rev{97.1} & \rev{95.6} & 97.8 & \rev{84.5} & \rev{\textbf{87.2}} \\
\bottomrule
\end{tabular}
}
\end{table*}
%--------------------------------------------------------------
\noindent\textbf{Results on NAVSIM v1.}
As shown in Table~\ref{tab:navsim_v1_main}, STEER achieves \textbf{91.2} PDMS on NAVSIM v1, \rev{competitive with state-of-the-art driving methods}. Crucially, while prior explicit adaptive methods like AdaThinkDrive~\cite{luo2025adathinkdrive} rely on human-annotated scene-difficulty labels to supervise routing, STEER achieves superior planning performance (91.2 vs. 90.3) \rev{without scene-complexity labels}. Furthermore, despite using a pure autoregressive VLM without an external planner, STEER \rev{numerically slightly} surpasses the planner-assisted baseline, SGDrive (91.1). Under Best-of-$N$ evaluation, STEER$^\dagger$ further reaches \textbf{93.4} PDMS, demonstrating exceptional candidate quality.

\noindent\textbf{Results on NAVSIM v2.}
As shown in Table~\ref{tab:navsim_v2}, STEER achieves the best \rev{\textbf{87.2}} EPDMS on NAVSIM v2. It improves over SGDrive (86.2) by \rev{\textbf{1.0}} points and also obtains the best DAC (\textbf{97.6}) and EP (\textbf{95.6}), indicating strong decision quality under the extended evaluation protocol.

Overall, STEER consistently improves planning quality across both benchmarks. As shown in the following routing analyses, these gains are accompanied by improved route selection quality, indicating that the model benefits from learning \emph{when} to invoke explicit reasoning rather than uniformly increasing reasoning length.

\subsection{Adaptive routing analysis}

%--------------------------------------------------------------
\begin{table*} [!t]
\centering
\caption{\textbf{Adaptive routing versus fixed routing on NAVSIM v1.} Adaptive routing is compared against forced direct and forced reasoning policies after route-balanced SFT and after RL.}
\label{tab:adaptive_routing}
\resizebox{\textwidth}{!}{
\begin{tabular}{l l c c c c c c c}
\toprule
Training Stage & Route Policy & NC$\uparrow$ & DAC$\uparrow$ & EP$\uparrow$ & TTC$\uparrow$ & C$\uparrow$ & PDMS$\uparrow$ & Avg. Resp. Tokens$\downarrow$ \\
\midrule
SFT
& Forced direct & 97.7 & 94.5 & 80.2 & 93.4 & 99.9 & 85.3 & 155 \\
& Forced reasoning  & 97.7 & 94.4 & 80.1 & 93.7 & 99.9 & 86.4 & 417 \\
& Adaptive      & 97.9 & 94.4 & 80.3 & 94.2 & 100.0 & \textbf{86.5} & 274 \\
\midrule
RL
& Forced direct & 97.6 & 96.5 & 89.0 & 92.1 & 99.9 & 89.9 & 156 \\
& Forced reasoning  & 97.5 & 97.0 & 90.3 & 91.9 & 100.0 & 90.8 & 415 \\
& Adaptive      & 97.4 & 97.3 & 90.2 & 92.3 & 100.0 & \textbf{91.2} & 315 \\
\bottomrule
\end{tabular}
}
\end{table*}
%--------------------------------------------------------------
Table~\ref{tab:adaptive_routing} compares STEER's learned adaptive routing with forced-direct and forced-reasoning policies. The results directly validate our core motivation regarding the pitfall of overthinking. After route-balanced SFT, the adaptive policy yields only a marginal gain over the fixed policies, indicating that this stage primarily preserves dual-route support rather than learning a strong routing preference. 

After RL, however, the forced-reasoning policy generates excessively long responses (415 tokens) but achieves a sub-optimal PDMS of 90.8. This confirms that blindly applying CoT to all scenes induces overthinking, which not only incurs unnecessary compute overhead but can also degrade planning quality. Importantly, the gain does not come from simply suppressing reasoning altogether: adaptive routing also outperforms the forced-direct policy (91.2 vs.\ 89.9), showing that STEER learns to preserve explicit reasoning precisely for scenes where it is beneficial. As a result, STEER achieves the highest PDMS of \textbf{91.2} while reducing the average response length to 315 tokens, demonstrating a superior effectiveness-efficiency trade-off.

%--------------------------------------------------------------
% \begin{figure} [!t]
%     \centering
%     \includesvg[width=1\linewidth]{figures/compare_0504}
%     \caption{\textbf{Prompt-level route comparison.}
%     For each prompt, we compare the trajectory produced by the route selected by STEER (blue) with the counterfactual alternative route (red). In the top case, explicit reasoning yields safer obstacle clearance. In the bottom case, STEER correctly avoids over-reasoning and selects the direct route, which better follows the road geometry compared to the perturbed reasoning alternative.}
%     \label{fig:think_answer_compare}
% \end{figure}
\begin{figure}[!t]
    \centering
    \includegraphics[width=1\linewidth]{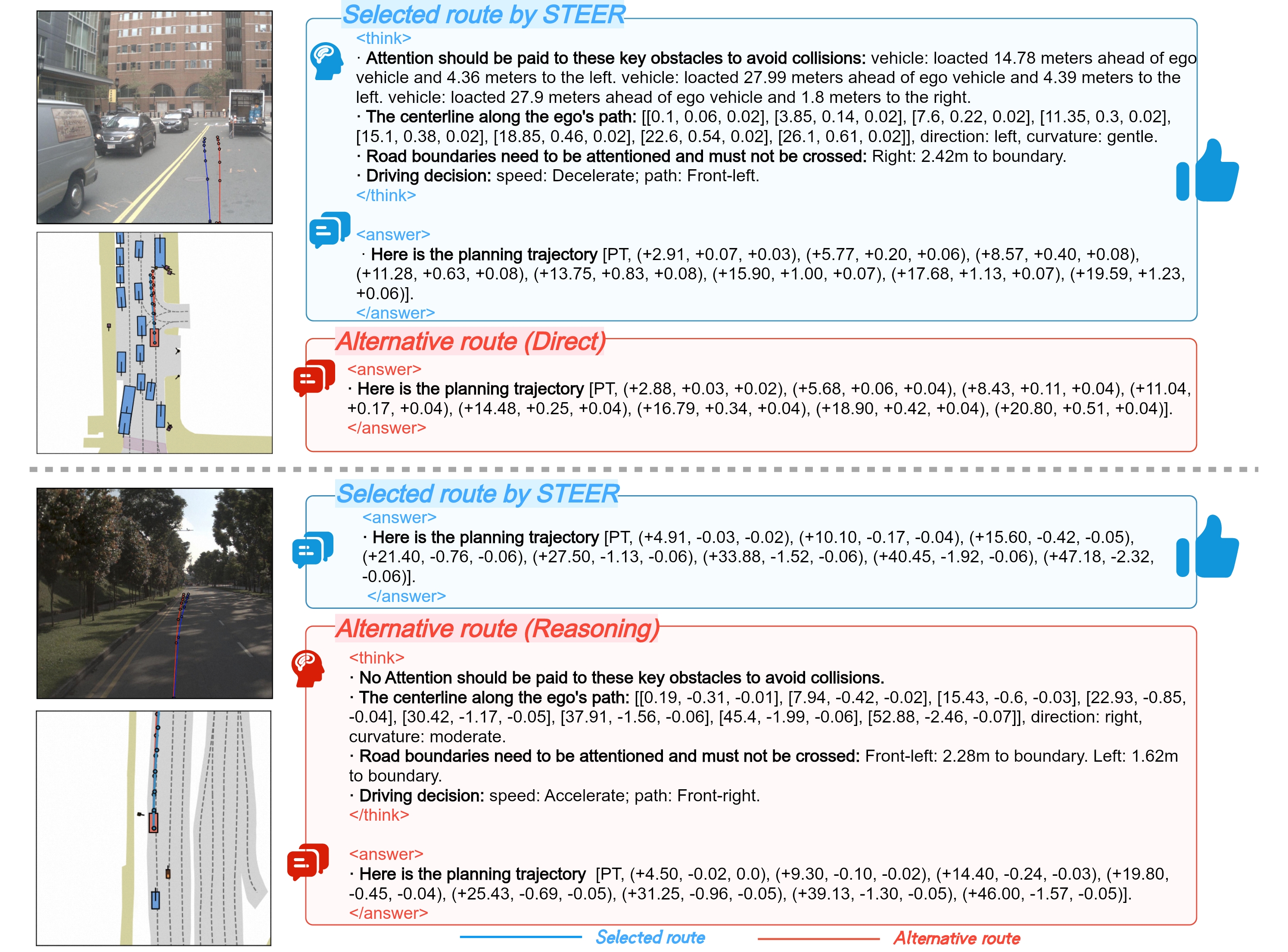}
    \caption{\textbf{Prompt-level route comparison.}
    For each prompt, we compare the trajectory produced by the route selected by STEER (blue) with the counterfactual alternative route (red). In the top case, explicit reasoning yields safer obstacle clearance. In the bottom case, STEER correctly avoids over-reasoning and selects the direct route, which better follows the road geometry compared to the perturbed reasoning alternative.}
    \label{fig:think_answer_compare}
\end{figure}
%--------------------------------------------------------------

To further understand the learned routing behavior, Figure~\ref{fig:think_answer_compare} shows prompt-level comparisons between the route selected by STEER and the counterfactual alternative for the same input. The examples suggest that the model does not uniformly favor either response mode. Instead, it invokes explicit reasoning when doing so improves safety or constraint handling, and correctly avoids over-reasoning by choosing direct generation when it is sufficient to produce a cleaner trajectory. Additional qualitative comparisons are provided in Appendix~\ref{app:qualitative}.

\subsection{Ablation studies}
\label{sec:ablations}

% --- RL Ablation Table ---
\begin{table*} [!t]
\centering
\caption{\textbf{Ablation of route-aware policy optimization}. Our proposed mechanisms effectively overcome exploration collapse and confounded credit assignment, achieving the best routing quality and planning performance.}
\label{tab:rl_ablation}
\setlength{\tabcolsep}{5pt}
\renewcommand{\arraystretch}{1.1}
\resizebox{\linewidth}{!}{%
\begin{tabular}{ccc|ccccc}
\toprule
\multirow{2}*{ID} & Routing Uncert.- & Cross-Route & \multirow{2}*{RouteAcc (\%)$\uparrow$} & \multirow{2}*{AUC (\%)$\uparrow$} & \multirow{2}*{AvgRegret$\downarrow$} & \multirow{2}*{Avg. Resp. Tokens$\downarrow$} & \multirow{2}*{PDMS$\uparrow$} \\
 & Aware Rollout & Advantage &  &  &  &  &  \\
\midrule
1 & $\times$   & $\times$   & \rev{69.3} & 73.9 & \rev{0.034} & \rev{363} & \rev{89.5} \\
2 & \checkmark & $\times$   & 70.7 & 74.4 & 0.023 & 362 & 90.9 \\
3 & $\times$   & \checkmark & 77.5 & 76.5 & 0.024 & 350 & 89.9 \\
4 & \checkmark & \checkmark & \rev{\textbf{81.1}} & \rev{\textbf{79.1}} & \textbf{0.018} & \rev{\textbf{315}} & \textbf{91.2} \\
\bottomrule
\end{tabular}%
}
\end{table*}
% -------------------------------------------------------------------

% --- SFT ablation ---
\begin{table}[!t]
\centering
\begin{revision}
\caption{\textbf{Ablation of route-balanced SFT.} Length normalization and routing calibration alleviate the direct-route bias.}
\label{tab:sft_ablation}
\small
\setlength{\tabcolsep}{6pt}
\renewcommand{\arraystretch}{1.1}
\resizebox{0.80\textwidth}{!}{%
\begin{tabular}{cccccc}
\toprule
ID & Length Norm. & Route Calib. & Reasoning Prob. (\%) & Direct Prob. (\%) & PDMS$\uparrow$ \\
\midrule
1 & $\times$ & $\times$ & 26.6 & 73.4 & 86.4 \\
2 & \checkmark & $\times$ & 43.2 & 56.8 & 86.1 \\
3 & \checkmark & \checkmark & \textbf{49.1} & \textbf{50.9} & \textbf{86.5} \\
\bottomrule
\end{tabular}%
}
\end{revision}
\end{table}

% --- Training pipeline ablation ---
\begin{table}[!t]
\centering
\begin{revision}
\caption{\textbf{Ablation of the training pipeline.} Planning performance with and without driving-centric pretraining and RL.}
\label{tab:stage_ablation}
\footnotesize
\setlength{\tabcolsep}{6pt}
\renewcommand{\arraystretch}{1.1}
\begin{tabular*}{0.95\textwidth}{@{\extracolsep{\fill}}ccc|cccccc}
\toprule
ID & Pretraining & RL & NC$\uparrow$ & DAC$\uparrow$ & EP$\uparrow$ & TTC$\uparrow$ & C$\uparrow$ & PDMS$\uparrow$ \\
\midrule
1 & $\times$ & $\times$ & 97.8 & 94.1 & 79.9 & 94.0 & 100.0 & 85.9 \\
2 & \checkmark & $\times$ & \textbf{97.9} & 94.4 & 80.3 & \textbf{94.2} & 100.0 & 86.5 \\
3 & \checkmark & \checkmark & 97.4 & 97.3 & 90.2 & 92.3 & 100.0 & 91.2 \\
\rev{4} & \rev{$\times$} & \rev{\checkmark} & \rev{97.5} & \rev{97.3} & \rev{90.2} & \rev{92.5} & \rev{100.0} & \rev{91.2} \\
\bottomrule
\end{tabular*}
\end{revision}
\end{table}

\begin{table}[!t]
\centering
\begin{revision}
\caption{\textbf{Rollout-mode dynamics during route-aware policy optimization.} Mode columns give percentages of prompts.}
\label{tab:rollout_dynamics}
\small
\setlength{\tabcolsep}{5pt}
\begin{tabular*}{\linewidth}{@{\extracolsep{\fill}}lcccc@{}}
\toprule
Point & Mode A & Mode B & Mode C & Reasoning ratio \\
\midrule
Start & 85\% & 13\% & 2\% & 50\% \\
Middle & 44\% & 35\% & 21\% & 57\% \\
End & 17\% & 22\% & 61\% & 64\% \\
\bottomrule
\end{tabular*}
\end{revision}
\end{table}

\paragraph{Effect of route-aware policy optimization.}
Table~\ref{tab:rl_ablation} isolates the contributions of our RL mechanisms. Vanilla GRPO (ID 1) improves over the route-balanced SFT initialization, but its routing regret remains relatively high, suggesting that sequence-level reward optimization alone does not reliably recover the latent route preference. Removing \textit{cross-route advantage credit assignment} (ID 2) leads to credit confounding between route selection and route-conditioned execution, making the model less effective at penalizing unnecessary reasoning, as reflected by the inflated response length (362 tokens) and lower PDMS (90.9). Removing \textit{routing uncertainty-aware rollout} (ID 3) deprives the model of counterfactual route comparison in uncertain scenes, causing exploration collapse and reducing PDMS to 89.9. Combining both components (ID 4) yields the best routing quality, the lowest regret, and the strongest planning performance, allowing STEER to achieve a \rev{competitive} PDMS of 91.2 entirely without scene-difficulty labels.

\paragraph{Effect of the route-balanced SFT.} \rev{In Table~\ref{tab:sft_ablation}, length normalization raises the reasoning probability from 26.6\% to 43.2\%, alleviating the direct-route bias. Routing calibration further balances the prior (49.1\%/50.9\%). With comparable PDMS (86.1--86.5), this prior supports subsequent reward-driven route learning.}

\begin{table}[!t]
\centering
\begin{revision}
\caption{\textbf{Efficiency analysis.} End-to-end inference latency (s/sample, left) and RL training cost (GPU-hours, right).}
\label{tab:efficiency}
\small
\setlength{\tabcolsep}{3pt}
\renewcommand{\arraystretch}{1.12}
\begin{minipage}[t]{0.49\linewidth}
\centering
\textbf{(a) Inference}\par\vspace{4pt}
\begin{tabular*}{\linewidth}{@{\extracolsep{\fill}}lcc@{}}
\toprule
Route policy & PDMS$\uparrow$ & Time (s/sample)$\downarrow$ \\
\midrule
Forced direct & 89.9 & \textbf{2.4} \\
Forced reasoning & 90.8 & 6.7 \\
STEER & \textbf{91.2} & 5.1 \\
\bottomrule
\end{tabular*}
\end{minipage}\hfill
\begin{minipage}[t]{0.49\linewidth}
\centering
\textbf{(b) RL training}\par\vspace{4pt}
\begin{tabular*}{\linewidth}{@{\extracolsep{\fill}}lc@{}}
\toprule
Training configuration & GPU-hours$\downarrow$ \\
\midrule
Vanilla GRPO & 2,560 \\
STEER w/o pretraining & 2,387 \\
STEER & \textbf{2,304} \\
\bottomrule
\end{tabular*}
\end{minipage}
\end{revision}
\end{table}

\paragraph{Effect of the training pipeline.}
Table~\ref{tab:stage_ablation} shows the contributions of the training stages. The SFT-only baseline (ID 1) achieves a PDMS of 85.9. Adding driving-centric pretraining (ID 2) provides a stronger driving-aware initialization, improving PDMS to 86.5. The largest gain then comes from route-aware policy optimization (ID 3), which further boosts PDMS to \rev{91.18}. \rev{Without pretraining, SFT followed by the same route-aware RL obtains 91.20 (ID 4). With comparable final PDMS, pretraining reduces RL-stage training cost from 2,387 to 2,304 GPU-hours, a 3.5\% reduction (Table~\ref{tab:efficiency}(b)).}

\paragraph{Rollout-mode dynamics.} \rev{The uncertainty-aware scheduler forces both routes for uncertain prompts (Modes A/B) to compare their task rewards, and uses more on-policy rollouts (Mode C) as routing confidence grows. Table~\ref{tab:rollout_dynamics} shows Mode A decreasing from 85\% to 17\% and Mode C increasing from 2\% to 61\% during training, while the budget stays at sixteen rollouts per prompt. At convergence, Modes A/B still cover 39\% of prompts, averaging 4.48 forced rollouts per prompt; the reasoning/direct ratio remains 64\%/36\%. Thus uncertain prompts continue to receive cross-route comparisons, while confident prompts use more of the budget to optimize the current policy and both routes remain represented overall.}

\subsection{\rev{Efficiency analysis}}
\label{sec:efficiency}

\begin{revision}
For inference, we measure end-to-end latency from model input to complete trajectory output over the full NAVSIM v1 navtest set on a single H20 GPU with batch size one and identical input resolution and decoding settings. Latencies are averaged over ten evaluation runs. Table~\ref{tab:efficiency}(a) uses the same checkpoint and route policies as Table~\ref{tab:adaptive_routing}. Relative to forced reasoning, STEER reduces latency from 6.7 to 5.1 s/sample (23.9\%) while improving PDMS from 90.8 to 91.2.

Table~\ref{tab:efficiency}(b) reports GPU-hours, calculated as $64\times$ the elapsed RL training time in hours, including rollout generation and forced-route allocation. STEER uses 10.0\% fewer GPU-hours than vanilla GRPO; pretraining reduces STEER's RL-stage cost by 3.5\%.
\end{revision}

\section{Conclusion}

In this work, we identified \emph{overthinking} as a critical cognitive pitfall in VLA-based autonomous driving. This led us to explore a central question: can adaptive reasoning with explicit route optimization be achieved entirely without subjective scene-difficulty labels? Our results \rev{support this feasibility claim}. We introduced \textbf{STEER}, a generic over-reasoning mitigation strategy \rev{without scene-complexity labels} that learns to balance CoT and direct responding directly from environmental rewards. By integrating \textit{route-balanced supervised fine-tuning} to preserve a dual-route prior, alongside \textit{routing uncertainty-aware rollout} and \textit{cross-route advantage credit assignment}, STEER successfully reinforces \rev{scene-dependent} routing decisions based on comparative task outcomes rather than \rev{manual difficulty annotations}. Extensive experiments on NAVSIM v1 and v2 demonstrate that STEER achieves \rev{planning performance competitive with state-of-the-art methods} while pushing the Pareto front of inference efficiency.
\begin{revision}
\renewcommand{\acksection}{\section*{Acknowledgments}}
\begin{ack}
This work was supported in part by New Generation Artificial Intelligence- National Science and Technology Major Project (2025ZD0123004), National Natural Science Foundation of China (Grant No.62376060) and Ningbo grant (2025Z038).
\end{ack}
\end{revision}
%%%%%%%%%%%%%%%%%%%%%%%%%%%%%%%%%%%%%%%%%%%%%%%%%%%%%%%%%%%%

\clearpage
\bibliographystyle{plainnat}
\bibliography{Reference}

\clearpage
\appendix

\section{Appendix overview}
\label{app:overview}

This appendix provides additional details that complement the main paper. We describe the data preparation pipeline, reward design, and full implementation details, and then report metric definitions, hyper-parameter settings, \rev{the route-surrogate analysis and controlled experiments,} sensitivity analyses, and additional qualitative visualizations.

\section{Data preparation}
\label{appendix:data_preparation}

This appendix summarizes the data preparation pipeline used in STEER. We construct two types of data: (1) \textbf{driving-centric pretraining data}, which adapts a general-purpose VLM to driving semantics and planning-relevant reasoning patterns, and (2) \textbf{dual-format SFT data}, which provides both reasoning-route and direct-route supervision under the same prompt distribution.

\subsection{Driving-centric pretraining data}
\label{app:pretraining_data}

The pretraining stage bridges general vision--language capability and autonomous-driving decision making before end-to-end trajectory generation. To this end, we construct three categories of driving-centric supervision.

\paragraph{Scene structure and key-obstacle understanding.}
We build instruction--answer pairs for lane topology recognition, drivable-area understanding, road-boundary identification, ego-lane localization, and key-obstacle identification under the current driving context. These tasks encourage the model to form structured scene representations and focus on decision-relevant agents.

\paragraph{Trajectory feasibility and safety judgment.}
Inspired by the trajectory-judgment formulation in MTDrive~\cite{li2026mtdrive}, we construct samples in which the model is given candidate future trajectories and asked to judge whether each trajectory is safe and feasible. This trains the model to connect scene geometry and agent interactions with future safety outcomes.

\paragraph{Public driving VLM datasets.}
We further incorporate public driving QA and reasoning datasets, including ReCogDrive~\cite{li2025recogdrive}, LingoQA~\cite{marcu2024lingoqa}, NuScenes-QA~\cite{qian2024nuscenes}, DriveLM~\cite{sima2024drivelm}, and OmniDrive~\cite{wang2025omnidrive}, to enrich semantic coverage and improve generalization.

\subsection{Dual-format SFT data}

To support explicit adaptive routing, each SFT sample is paired with two valid targets under the same prompt: a \textbf{direct route} target and a \textbf{reasoning route} target. Each sample contains a front-view image, structured driving context, and the target future ego trajectory. 

For the reasoning route, we use a structured CoT format aligned with planning-relevant driving cues. Each reasoning trace contains four components: (1) identifying the key obstacle under the current scene context, (2) describing the road centerline or nominal drivable path, (3) analyzing road layout and boundary constraints, and (4) summarizing a high-level driving decision, including lateral intent and longitudinal speed trend. This structure encourages reasoning traces that remain closely aligned with downstream planning.

\subsection{Automatic annotation pipeline}

\begin{figure}[!t]
    \centering
    \includegraphics[width=1\linewidth]{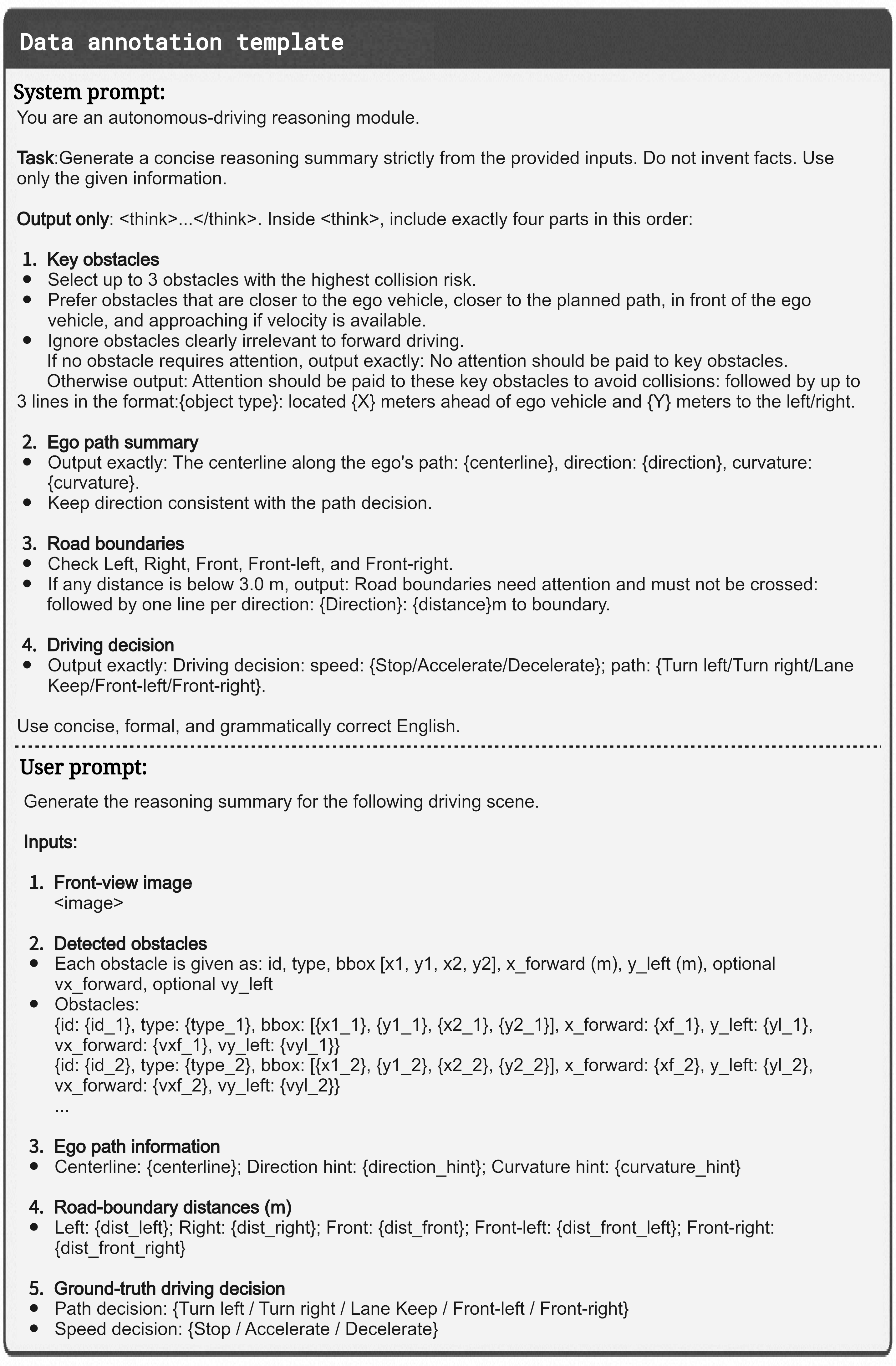}
    \caption{Prompt template for reasoning-data generation.}
    \label{fig:anno_template}
\end{figure}

We construct the above supervision with an automatic annotation pipeline that combines rule-based grounding and VLM-based distillation. Objective scene elements such as lane geometry, road boundaries, ego-lane position, dynamic object states, trajectory targets, and navigation signals are first extracted from structured annotations and simulator metadata. These grounded signals are then organized into a prompt template together with the front-view image and fed to a strong teacher VLM (Qwen2.5-VL-72B~\cite{bai2025qwen25vltechnicalreport}) to synthesize the reasoning trace. Figure~\ref{fig:anno_template} illustrates the prompt template used for reasoning-data generation.

The prompt template is designed to expose the teacher model to the core factors relevant to planning. On the input side, it provides static-road information (e.g., road boundaries and drivable-area geometry), dynamic-object attributes (e.g., position, heading, speed, and object class), and ego-state information (e.g., ego pose, motion history, navigation instruction, and front-view image). On the output side, the teacher model is asked to generate planning-oriented reasoning rather than free-form scene descriptions. In particular, the generated CoT is required to cover the same four components used in our reasoning-route supervision: key-obstacle identification, road centerline or nominal path, road layout and boundary analysis, and high-level driving decision.

Within this pipeline, rule-based processing is used to provide reliable scene grounding, while the teacher VLM is responsible for converting these grounded signals into coherent natural-language reasoning. For example, road-boundary distances and drivable-area geometry can be computed directly from structured map annotations, while key-obstacle selection and high-level decision description are distilled from the teacher model under grounded prompts. This hybrid design combines the factual precision of structured annotations with the flexibility of language generation.

These data components provide driving-aware initialization and dual-route execution supervision, but they do not provide route-preference labels or scene-difficulty annotations. The final routing policy is learned later from environmental rewards during route-aware policy optimization.
\rev{The 72B teacher constructs driving data and CoT execution targets offline, without assessing scene complexity or providing routing labels.}

\subsection{Training data template}

\begin{figure}[!t]
    \centering
    \includegraphics[width=1\linewidth]{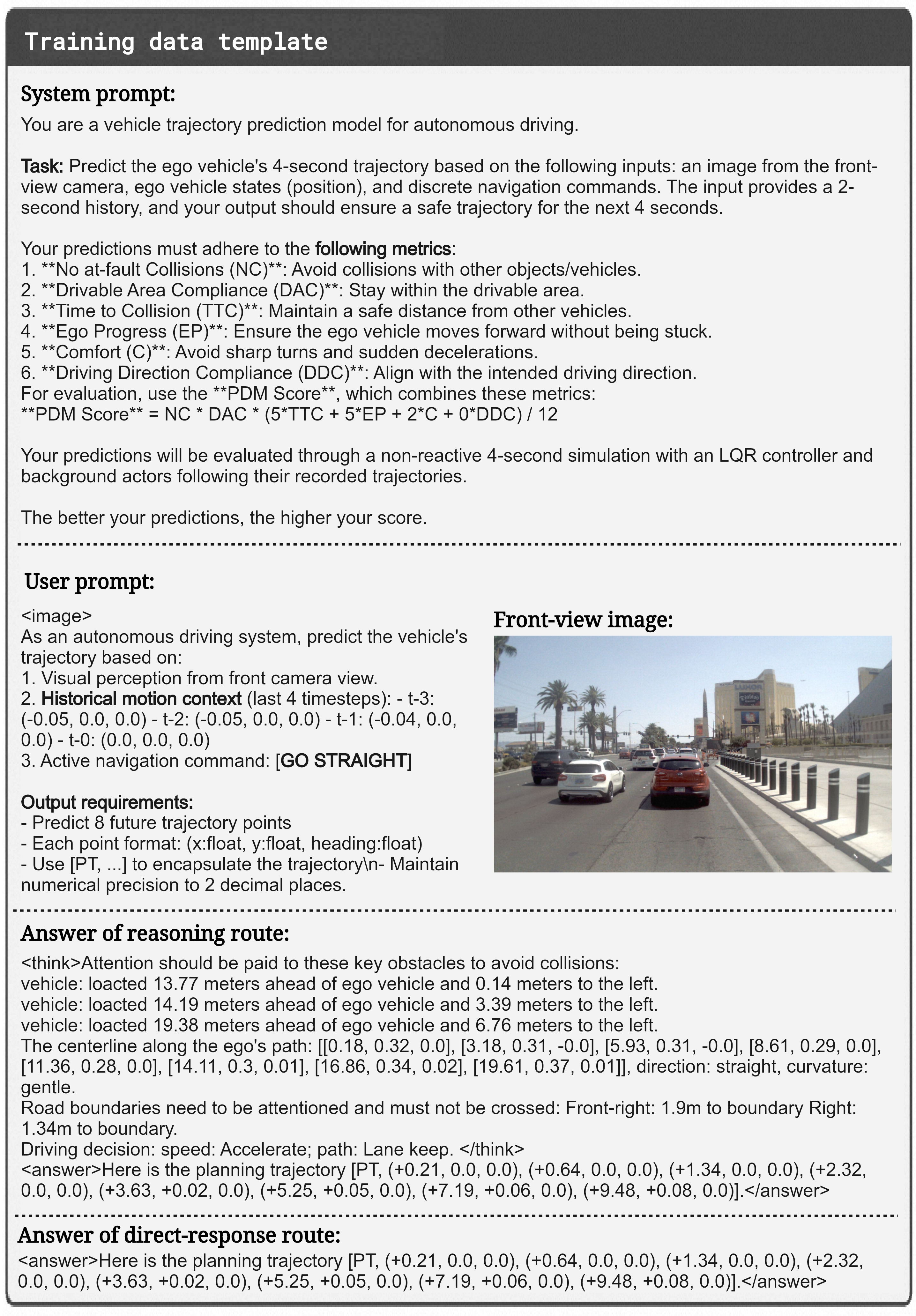}
    \caption{Training data template used for driving-centric pretraining and SFT.}
    \label{fig:data_prompt}
\end{figure}

Figure~\ref{fig:data_prompt} shows an example of the training data template used in our experiments, including the task description, sensory inputs, and target trajectory format.

\section{Driving-centric pretraining details}
\label{app:pretrain}

The driving-centric pretraining stage is used to adapt the base vision-language model to autonomous-driving tasks before route-balanced SFT and route-aware policy optimization. We use a driving-specific pretraining corpus constructed from multiple data sources and automatic annotation pipelines. Detailed data composition and construction are provided in Appendix~\ref{app:pretraining_data}.

\section{Reward design}
\label{appendix:reward}
This appendix details the scalar reward used in the RL stage of STEER. The reward is designed to supervise response validity, trajectory quality, and reasoning-action consistency. Importantly, the reward itself does not impose any \rev{scene-complexity or route-choice label}. The actual learning signal for route selection is provided by the cross-route advantage defined in Section~\ref{sec:route_aware_po}, which compares the realized outcomes of the reasoning and direct routes under the same prompt.

\paragraph{Format reward.}
A binary reward $r^{\texttt{fmt}} \in \{0, 1\}$ checks whether the output follows one of the two valid response formats, namely the direct route or the reasoning route. For the reasoning route, we additionally require that the \texttt{<think>} block does not contain nested \texttt{<answer>} tags. If the format check fails, the rollout receives zero effective task reward.

\paragraph{Planning reward.}
We use PDMS~\cite{dauner2024navsim} as the main trajectory-quality signal:
\begin{equation}
r^{\texttt{pdms}} = \mathrm{PDMS}(\hat{\boldsymbol{\tau}}),
\end{equation}
where $\hat{\boldsymbol{\tau}}$ is the predicted trajectory parsed from the \texttt{<answer>} block. PDMS jointly evaluates safety, drivable-area compliance, and route progress, and serves as the dominant task objective during RL.

\paragraph{Reasoning-action consistency reward.}
For the reasoning route, we introduce a binary consistency reward $r^{\texttt{cons}} \in \{0, 1\}$ that measures whether the high-level decision expressed in \texttt{<think>} is consistent with the generated trajectory. Concretely, we compare two aspects: speed trend and path intent. The reward is defined as
\begin{equation}
\label{eq:consistency_appendix}
r^{\texttt{cons}} =
\mathbb{1}\!\left[
s_{\texttt{reason}} = s_{\texttt{traj}}
\;\land\;
p_{\texttt{reason}} = p_{\texttt{traj}}
\right],
\end{equation}
where $s$ and $p$ denote the speed and path categories, respectively. This term acts as a lightweight regularizer that encourages the generated reasoning to remain aligned with the final planning output, rather than drifting into irrelevant or contradictory deliberation.

\paragraph{Final scalar reward.}
We use a route-dependent scalar reward:
\begin{equation}
\label{eq:base_reward_appendix}
r =
\begin{cases}
\alpha_f\, r^{\texttt{fmt}} + \alpha_p\, r^{\texttt{pdms}} + \alpha_c\, r^{\texttt{cons}}, & \text{if the reasoning route is used}, \\[4pt]
\alpha_f\, r^{\texttt{fmt}} + (1-\alpha_f)\, r^{\texttt{pdms}}, & \text{if the direct route is used}.
\end{cases}
\end{equation}
Here, $r^{\texttt{pdms}}$ remains the dominant signal in both routes. The consistency term is only applied to the reasoning route, where explicit deliberation is present, and serves as a mild auxiliary regularizer rather than a primary optimization target.
\rev{With the weights in Table~\ref{tab:hyperparams}, the direct reward is $0.9r^{\texttt{pdms}}+0.1r^{\texttt{fmt}}$, whereas the reasoning reward is $0.8r^{\texttt{pdms}}+0.1r^{\texttt{fmt}}+0.1r^{\texttt{cons}}$. Both weight sums equal one, normalizing the maximum reward scale across routes. The consistency term contributes to the scalar return used for route learning.}

Overall, this reward formulation provides task-level supervision for valid response generation and driving performance, \rev{with comparative policy optimization learning which route to prefer from these total returns, rather than from routing annotations}.

\section{Experimental details}
\label{app:exp_details}

\subsection{Detailed metric definitions}
\label{app:metric}

In the main paper, we report driving planning metrics from the NAVSIM benchmark~\cite{dauner2024navsim} as well as custom routing metrics (\textbf{RouteAcc}, \textbf{AUC}, and \textbf{AvgRegret}) to evaluate routing quality. Here we provide their full definitions.

\paragraph{Driving planning metrics.}
Following the NAVSIM benchmark, we evaluate trajectory quality using the \rev{Predictive Driver Model Score} (PDMS) for v1 and \rev{Extended Predictive Driver Model Score} (EPDMS) for v2.

\textbf{NAVSIM v1 Metrics:} PDMS is a composite metric that heavily weights the driving progress while strictly penalizing safety and comfort violations. It is computed based on several sub-metrics:
\begin{itemize}[leftmargin=*]
    \item \textbf{NC (No Collision):} \rev{A severity-aware at-fault collision score taking values in $\{0,0.5,1\}$; its mean is not a binary collision-event rate.}
    \item \textbf{DAC (Drivable Area Compliance):} Evaluates whether the trajectory stays within the designated drivable area. \rev{This is a binary per-scenario compliance score.}
    \item \textbf{EP (Ego Progress):} \rev{Measures the ego vehicle's normalized longitudinal progress along the reference route.}
    \item \textbf{TTC (Time to Collision):} \rev{A binary per-scenario score checking prospective collision risk; reported TTC scores are compliance percentages, rather than times in seconds.}
    \item \textbf{C (Comfort):} Penalizes excessive acceleration, deceleration, or jerk to ensure a smooth ride.
\end{itemize}

\textbf{NAVSIM v2 Metrics:} EPDMS extends PDMS by incorporating additional fine-grained compliance checks:
\begin{itemize}[leftmargin=*]
    \item \rev{\textbf{DDC (Driving Direction Compliance):} Penalizes driving against the permitted traffic direction.}
    \item \textbf{TLC (Traffic Light Compliance):} Checks adherence to traffic light signals.
    \item \textbf{LK (Lane Keeping):} Evaluates lateral stability and alignment with the lane.
    \item \rev{\textbf{HC (History Comfort):} Evaluates comfort using the historical and predicted ego motion.}
    \item \rev{\textbf{EC (Extended Comfort):} Evaluates the consistency of predicted accelerations and jerks across consecutive frames.}
\end{itemize}

\paragraph{Routing quality metrics.}
For each prompt $\mathbf{x}$, we force the model to follow the reasoning route and the direct-response route, obtaining PDMS scores $r^{\texttt{reason}}(\mathbf{x})$ and $r^{\texttt{direct}}(\mathbf{x})$, respectively. Let $\hat{a}(\mathbf{x}) \in \{\texttt{reason}, \texttt{direct}\}$ denote the route selected under free decoding.

\textbf{RouteAcc} measures whether the selected route matches the higher-reward route:
\begin{equation}
\text{RouteAcc}=
\frac{1}{N}\sum_{\mathbf{x}}
\mathbb{1}\!\left[
\hat{a}(\mathbf{x})=
\arg\max_{a\in\{\texttt{reason},\texttt{direct}\}} r^a(\mathbf{x})
\right].
\end{equation}

\textbf{AUC} measures how well the predicted reasoning-route probability ranks prompts according to whether the reasoning route is preferable. For each prompt, we define a binary label
\begin{equation}
y(\mathbf{x}) = \mathbb{1}\!\left[r^{\texttt{reason}}(\mathbf{x}) > r^{\texttt{direct}}(\mathbf{x})\right],
\end{equation}
and use the predicted reasoning-route probability $P_{\texttt{reason}}(\mathbf{x})$ as the ranking score. Higher AUC indicates that the model assigns larger reasoning probabilities to prompts for which the reasoning route is indeed more beneficial.

\textbf{AvgRegret} measures the gap between the selected route and the oracle route:
\begin{equation}
\text{AvgRegret}=
\frac{1}{N}\sum_{\mathbf{x}}
\left(
\max\{r^{\texttt{reason}}(\mathbf{x}),\,r^{\texttt{direct}}(\mathbf{x})\}
-r^{\hat{a}(\mathbf{x})}(\mathbf{x})
\right).
\end{equation}

For the route-balanced SFT stage analysis, we additionally report routing-distribution statistics based on the initial decision probabilities. Let $P_{\texttt{reason}}(\mathbf{x})$ and $P_{\texttt{direct}}(\mathbf{x})$ denote the predicted probabilities of taking the reasoning and direct-response routes, respectively. We report their dataset averages,
\begin{equation}
\bar{P}_{\texttt{reason}}=\frac{1}{N}\sum_{\mathbf{x}} P_{\texttt{reason}}(\mathbf{x}), \qquad
\bar{P}_{\texttt{direct}}=\frac{1}{N}\sum_{\mathbf{x}} P_{\texttt{direct}}(\mathbf{x}),
\end{equation}
as well as the average routing uncertainty (entropy):
\begin{equation}
\text{RoutingUncertainty}
=
-\frac{1}{N}\sum_{\mathbf{x}}
\sum_{a\in\{\texttt{reason},\texttt{direct}\}}
P_a(\mathbf{x})\log P_a(\mathbf{x}).
\end{equation}

\subsection{Detailed implementation}
\label{app:impl}

\textbf{Model.}
STEER is initialized from Qwen2.5-VL-8B and directly auto-regresses text waypoint trajectories without an additional planner. Given the current front-view image and the ego vehicle's historical motion, the model predicts a 4-second future trajectory as 8 waypoint tokens in natural language format. The historical trajectory is represented as numerical coordinates and concatenated into the prompt.

\textbf{Training pipeline.}
Training consists of three stages: driving-centric pretraining, route-balanced SFT, and route-aware policy optimization. In pretraining, we use about 600K samples for 2 epochs with a global batch size of 256 and a learning rate of \(4.0\times10^{-5}\). In SFT, we train on 80K NAVSIM samples for 5 epochs with a global batch size of 256 and a learning rate of \(4.0\times10^{-5}\). In RL, we use 40K NAVSIM samples for 5 epochs with a global batch size of 256, a learning rate of \(3.0\times10^{-6}\), and 16 rollouts per prompt.

\textbf{Training setup.}
All stages use full-parameter fine-tuning and are implemented with DeepSpeed ZeRO-3 on 64 NVIDIA H20 GPUs. Additional system and optimization details follow the standard distributed training setup used in our codebase.

\section{Hyper-parameter settings}
\label{appendix:hyperparams}

We summarize the main hyper-parameter settings used in SFT and RL in Table~\ref{tab:hyperparams}. Unless otherwise specified, all experiments use the same backbone, input resolution, and distributed training setup.

\begin{table*}[!t]
\centering
\caption{\textbf{Main hyper-parameter settings for route-balanced SFT and route-aware policy optimization.}}
\label{tab:hyperparams}
\setlength{\tabcolsep}{8pt}
\renewcommand{\arraystretch}{1.15}
\resizebox{0.95\textwidth}{!}{%
\begin{tabular}{l c @{\hspace{20pt}} l c}
\toprule
\multicolumn{2}{c}{\textbf{SFT}} & \multicolumn{2}{c}{\textbf{Route-aware policy optimization}} \\
\cmidrule(r){1-2} \cmidrule(l){3-4}
Parameter & Value & Parameter & Value \\
\midrule
Base model                        & Qwen2.5-VL-8B        & Rollout group size $n$                & 16 \\
Max pixels                        & 2073600              & Clipping coefficient $\epsilon_c$     & 0.2 \\
Global batch size                 & 256                  & Minority-route threshold $\epsilon$   & 0.05 \\
Epochs                            & 5                    & Low-entropy threshold $\tau_{\text{low}}$ & 0.25 \\
Learning rate                     & \(4.0\times10^{-5}\) & High-entropy threshold $\tau_{\text{high}}$ & 0.50 \\
Weight decay                      & 0.05                 & Global batch size                     & 256 \\
Warmup ratio                      & 0.1                  & Epochs                                & 5 \\
bfloat16                          & Enabled              & Learning rate                         & \(3.0\times10^{-6}\) \\
Routing calibration weight $\lambda$ & 0.1              & Weight decay                          & 0.01 \\
Route-balance coefficient $\alpha$ & 0.5                & KL coefficient $\beta_{\mathrm{KL}}$  & 0.1 \\
GPUs                              & 64                   & Max response length                   & 1000 \\
                                  &                      & Route loss weight $\lambda_r$         & 0.05 \\
                                  &                      & Format reward weight $\alpha_f$       & 0.1 \\
                                  &                      & PDMS reward weight $\alpha_p$ (reasoning route) & 0.8 \\
                                  &                      & Consistency reward weight $\alpha_c$  & 0.1 \\
                                  &                      & PDMS reward weight $(1-\alpha_f)$ (direct route) & 0.9 \\
                                  &                      & GPUs                                  & 64 \\
\bottomrule
\end{tabular}%
}
\end{table*}

\section{\rev{Supplementary analysis and controlled experiments}}
\label{app:supplementary}

\subsection{\rev{Evaluation statistics}}
\label{app:eval_statistics}
\begin{revision}
Tables~\ref{tab:navsim_v1_statistics}--\ref{tab:pretraining_control} report mean $\pm$ standard deviation over three evaluation seeds for each listed fixed checkpoint. The vanilla-GRPO and STEER conditions in Table~\ref{tab:matched_grpo} share the same backbone, initialization, data, reward, split, sixteen-rollout budget, and optimization settings; they differ in uncertainty-aware rollout and cross-route advantage credit assignment.
\end{revision}

\begin{table}[!htbp]
\centering
\begin{revision}
\caption{\textbf{NAVSIM v1 evaluation statistics.} Mean $\pm$ standard deviation over three evaluation seeds.}
\label{tab:navsim_v1_statistics}
\small
\setlength{\tabcolsep}{4pt}
\resizebox{\linewidth}{!}{%
\begin{tabular}{lcccccc}
\toprule
Setting & NC$\uparrow$ & DAC$\uparrow$ & EP$\uparrow$ & TTC$\uparrow$ & C$\uparrow$ & PDMS$\uparrow$ \\
\midrule
STEER & $97.41 \pm 0.36$ & $97.25 \pm 0.55$ & $90.22 \pm 0.81$ & $92.25 \pm 1.10$ & $100.0 \pm 0.0$ & $91.18 \pm 0.25$ \\
STEER (Best-of-6) & $98.02 \pm 0.19$ & $97.73 \pm 0.31$ & $91.61 \pm 0.32$ & $94.36 \pm 0.51$ & $100.0 \pm 0.0$ & $93.44 \pm 0.12$ \\
\bottomrule
\end{tabular}%
}
\end{revision}
\end{table}

\begin{table}[!htbp]
\centering
\begin{revision}
\caption{\textbf{NAVSIM v2 evaluation statistics for STEER.} Mean $\pm$ standard deviation over three evaluation seeds.}
\label{tab:navsim_v2_statistics}
\small
\setlength{\tabcolsep}{12pt}
\begin{tabular}{lc|lc}
\toprule
Metric & Score & Metric & Score \\
\midrule
NC$\uparrow$ & $97.40 \pm 1.21$ & TTC$\uparrow$ & $97.12 \pm 0.92$ \\
DAC$\uparrow$ & $97.63 \pm 0.58$ & LK$\uparrow$ & $95.61 \pm 1.31$ \\
DDC$\uparrow$ & $98.05 \pm 1.10$ & HC$\uparrow$ & $97.81 \pm 0.08$ \\
TLC$\uparrow$ & $99.15 \pm 0.15$ & EC$\uparrow$ & $84.47 \pm 1.12$ \\
EP$\uparrow$ & $95.62 \pm 0.22$ & EPDMS$\uparrow$ & $87.15 \pm 0.33$ \\
\bottomrule
\end{tabular}
\end{revision}
\end{table}

\begin{table}[!htbp]
\centering
\begin{revision}
\caption{\textbf{Matched comparison with vanilla GRPO.} Mean $\pm$ standard deviation over three evaluation seeds for IDs 1 and 4 of Table~\ref{tab:rl_ablation}.}
\label{tab:matched_grpo}
\small
\setlength{\tabcolsep}{4pt}
\resizebox{\linewidth}{!}{%
\begin{tabular}{lccccc}
\toprule
Setting & RouteAcc (\%)$\uparrow$ & AUC (\%)$\uparrow$ & AvgRegret$\downarrow$ & Avg. tokens$\downarrow$ & PDMS$\uparrow$ \\
\midrule
Vanilla GRPO & $69.33 \pm 0.68$ & $73.91 \pm 0.31$ & $0.034 \pm 0.005$ & $363 \pm 6$ & $89.45 \pm 0.31$ \\
STEER & $81.09 \pm 0.34$ & $79.06 \pm 0.40$ & $0.018 \pm 0.002$ & $315 \pm 4$ & $91.18 \pm 0.25$ \\
\bottomrule
\end{tabular}%
}
\end{revision}
\end{table}

\begin{table}[!htbp]
\centering
\begin{revision}
\caption{\textbf{Pretraining ablation after route-aware RL.} Mean $\pm$ standard deviation over three evaluation seeds; both settings use route-balanced SFT and the same RL procedure.}
\label{tab:pretraining_control}
\small
\setlength{\tabcolsep}{4pt}
\resizebox{\linewidth}{!}{%
\begin{tabular}{lcccccc}
\toprule
Setting & NC$\uparrow$ & DAC$\uparrow$ & EP$\uparrow$ & TTC$\uparrow$ & C$\uparrow$ & PDMS$\uparrow$ \\
\midrule
With pretraining & $97.41 \pm 0.36$ & $97.25 \pm 0.55$ & $90.22 \pm 0.81$ & $92.25 \pm 1.10$ & $100.0 \pm 0.0$ & $91.18 \pm 0.25$ \\
Without pretraining & $97.45 \pm 0.40$ & $97.31 \pm 0.34$ & $90.17 \pm 0.63$ & $92.54 \pm 0.90$ & $100.0 \pm 0.0$ & $91.20 \pm 0.28$ \\
\bottomrule
\end{tabular}%
}
\end{revision}
\end{table}

\subsection{\rev{Expected-update analysis of the route surrogate}}
\label{app:route_analysis}
\begin{revision}
For a fixed prompt, let $\mu_{\mathrm r}=\mathbb{E}[r\mid\mathcal R=\texttt{reason}]$, $\mu_{\mathrm d}=\mathbb{E}[r\mid\mathcal R=\texttt{direct}]$, and $\Delta=\mu_{\mathrm r}-\mu_{\mathrm d}$, where $r$ is the total scalar reward. We analyze the additional route-specific update induced by $\mathcal{L}^{\texttt{route}}$ using the population form of Eq.~\ref{eq:route_advantage}, with both routes represented, holding route-conditioned returns fixed and ignoring clipping and KL regularization. The score update is evaluated at $\theta=\theta_{\mathrm{old}}$, where the token probability ratio equals one. For forced routes, this ratio still uses the old model probability; route allocation determines the sampling proportions below.

Writing $p=\pi_\theta(\mathcal R=\texttt{reason}\mid\mathbf x)=\sigma(z)$ and denoting the sampled route proportions by $q_{\mathrm r}$ and $q_{\mathrm d}$, with $q_{\mathrm r}+q_{\mathrm d}=1$, the population advantages satisfy
\begin{equation}
\mathbb{E}[A^{\texttt{route}}\mid\mathcal R=\texttt{reason}]=\Delta,
\qquad
\mathbb{E}[A^{\texttt{route}}\mid\mathcal R=\texttt{direct}]=-\Delta.
\label{eq:route_population}
\end{equation}
Since $\partial_z\log p=1-p$ and $\partial_z\log(1-p)=-p$, substituting these derivatives and Eq.~\ref{eq:route_population} gives the route-logit update in Eq.~\ref{eq:route_expected_update}. For $0<p<1$ with both routes represented, the coefficient is positive: the update favors reasoning when $\Delta>0$, favors direct responding when $\Delta<0$, and has zero expected cross-route signal when $\Delta=0$.

In the purely on-policy population case, $q_{\mathrm r}=p$ and $q_{\mathrm d}=1-p$, so Eq.~\ref{eq:route_expected_update} becomes $2p(1-p)\Delta$. The expected route reward and its logit gradient are
\begin{equation}
J(p)=p\mu_{\mathrm r}+(1-p)\mu_{\mathrm d},
\qquad
\partial_z J=p(1-p)\Delta.
\label{eq:route_reward_gradient}
\end{equation}
In this on-policy population setting, the two updates have the same direction and stationary conditions ($\Delta=0$ or limiting boundary policies). Forced-route proportions preserve the interior preference direction. This analysis applies to the simplified population update.

The per-rollout surrogate decomposes as
\begin{equation}
A_i^{\texttt{route}}=
\begin{cases}
\Delta+(r_i-\mu_{\mathrm r}), & \mathcal R_i=\texttt{reason},\\
-\Delta+(r_i-\mu_{\mathrm d}), & \mathcal R_i=\texttt{direct}.
\end{cases}
\label{eq:route_contrast_decomposition}
\end{equation}
The within-route residual has zero conditional expectation, so this form and a pure route-level contrast $\pm\Delta$ give the same expected preference direction under the stated assumptions. The per-rollout form integrates directly into the token-level clipped GRPO/PPO objective. An empirical comparison with pure route-level contrast remains future work.

The reward has no length or latency penalty: equal total expected rewards give zero expected cross-route update. Efficiency gains are empirical; a length penalty could reinforce the direct-route bias observed in vanilla SFT (Table~\ref{tab:sft_ablation}).
\end{revision}

\subsection{\rev{Safety and progress analysis}}
\label{app:safety}
\begin{revision}
In Table~\ref{tab:adaptive_routing}, the same RL checkpoint is evaluated under three route policies. For the binary DAC and TTC metrics, the violation rate is $100\%-\text{score}\%$. Forced direct, forced reasoning, and adaptive STEER have DAC-violation rates of 3.5\%, 3.0\%, and 2.9\%, and TTC-violation rates of 7.9\%, 8.1\%, and 7.7\%, respectively. Adaptive STEER has higher TTC, DAC, EP, and PDMS point estimates, while NC is 0.1--0.2 points lower. This controlled comparison does not show a TTC-for-progress trade-off. The cross-system results in Tables~\ref{tab:navsim_v1_main} and~\ref{tab:navsim_v2} reflect different safety/progress balances.

NC is severity-aware (Appendix~\ref{app:metric}). Collision-event counts are unavailable in the retained aggregate outputs, so a separate binary collision rate cannot be reconstructed.
\end{revision}

\section{Sensitivity analysis}
\label{appendix:sensitivity}

We further analyze the sensitivity of two key route-related hyper-parameters: the route calibration weight $\lambda$ in SFT and the route loss weight $\lambda_r$ in RL. Tables~\ref{tab:sensitivity_sft_lambda} and~\ref{tab:sensitivity_rl_lambda} report the results on NAVSIM v1.

\begin{table}[!t]
\centering
\caption{\textbf{Sensitivity analysis of the route-balanced SFT calibration weight $\lambda$.} Moderate route calibration improves route balance and yields the best downstream planning performance.}
\label{tab:sensitivity_sft_lambda}
\small
\setlength{\tabcolsep}{6pt}
\renewcommand{\arraystretch}{1.08}
\begin{tabular}{ccccc}
\toprule
$\lambda$ & Reasoning Prob. (\%) & Direct Prob. (\%) & Routing Uncertainty$\uparrow$ & PDMS$\uparrow$ \\
\midrule
0    & 43.2 & 56.8 & 0.681 & 86.1 \\
0.05 & 48.5 & 51.5 & 0.689 & 86.0 \\
0.10 & 49.1 & 50.9 & \textbf{0.693} & \textbf{86.5} \\
0.15 & 49.4 & 50.6 & 0.692 & 86.0 \\
0.20 & 49.6 & 50.4 & 0.692 & 85.9 \\
\bottomrule
\end{tabular}
\end{table}

\begin{table}[!t]
\centering
\caption{\textbf{Sensitivity analysis of the RL route loss weight $\lambda_r$.} A moderate route loss weight provides the best routing quality, efficiency, and planning performance.}
\label{tab:sensitivity_rl_lambda}
\small
\setlength{\tabcolsep}{5pt}
\renewcommand{\arraystretch}{1.08}
\begin{tabular}{cccccc}
\toprule
$\lambda_r$ & RouteAcc (\%)$\uparrow$ & AUC (\%)$\uparrow$ & AvgRegret$\downarrow$ & Avg. Resp. Tokens$\downarrow$ & PDMS$\uparrow$ \\
\midrule
0    & 70.7 & 74.4 & 0.023 & 362 & 90.9 \\
0.02 & 77.8 & 77.4 & 0.020 & 333 & 91.1 \\
0.05 & \textbf{81.1} & \textbf{79.1} & \textbf{0.018} & \textbf{315} & \textbf{91.2} \\
0.10 & 80.1 & 76.8 & 0.029 & 322 & 90.4 \\
0.20 & 77.2 & 74.7 & 0.039 & 324 & 89.4 \\
\bottomrule
\end{tabular}
\end{table}

\begin{figure} [!t]
    \centering
    \includegraphics[trim=140bp 4bp 140bp 480bp,clip,width=\linewidth]{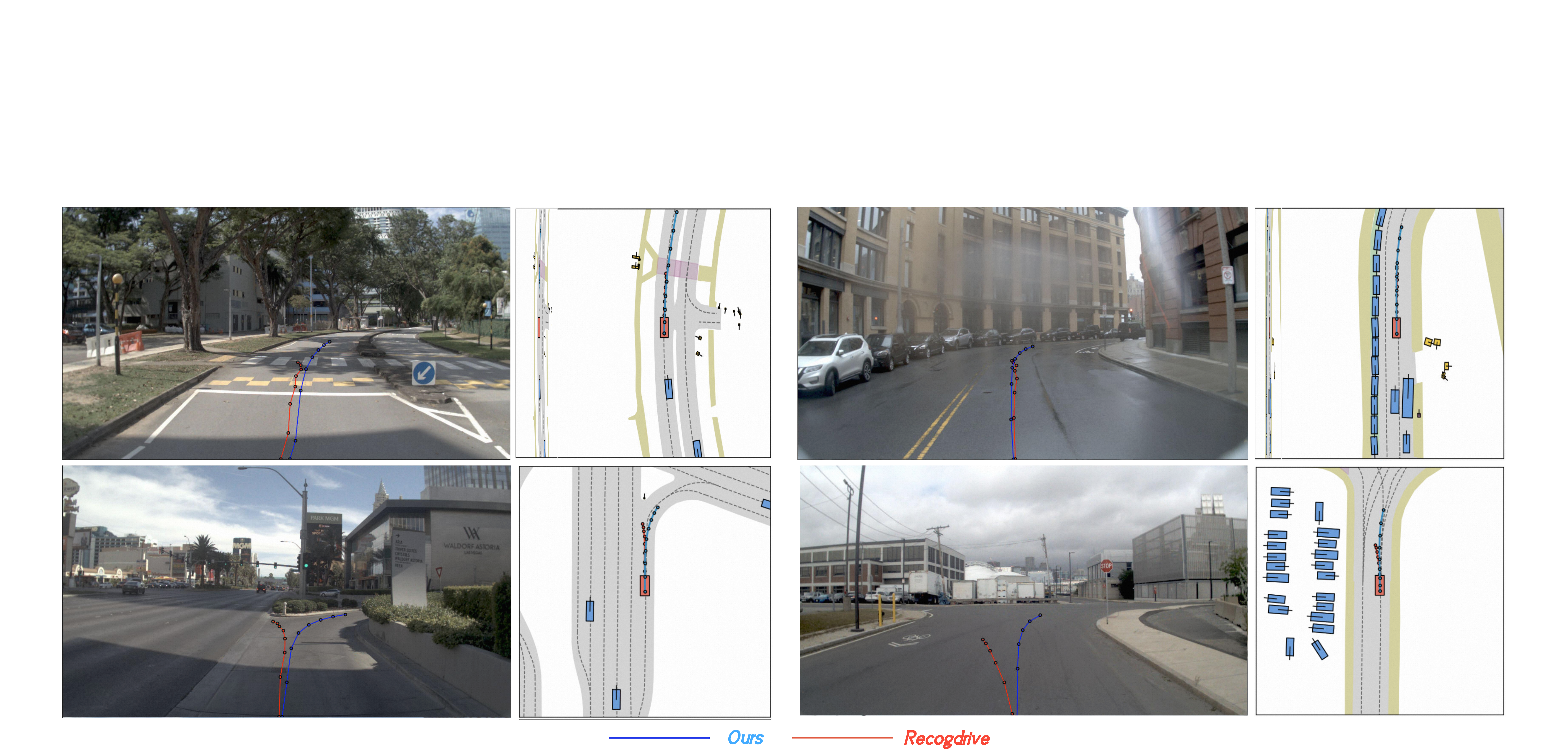}
    \caption{\textbf{Qualitative comparison with a representative baseline.}
    We compare STEER (blue) with ReCogDrive~\cite{li2025recogdrive} (red) on representative prompts. STEER produces trajectories that better align with the road geometry and intended maneuver.}
    \label{fig:qual_compare_app}
\end{figure}

\begin{figure}[!t]
    \centering
    \includegraphics[width=1\linewidth]{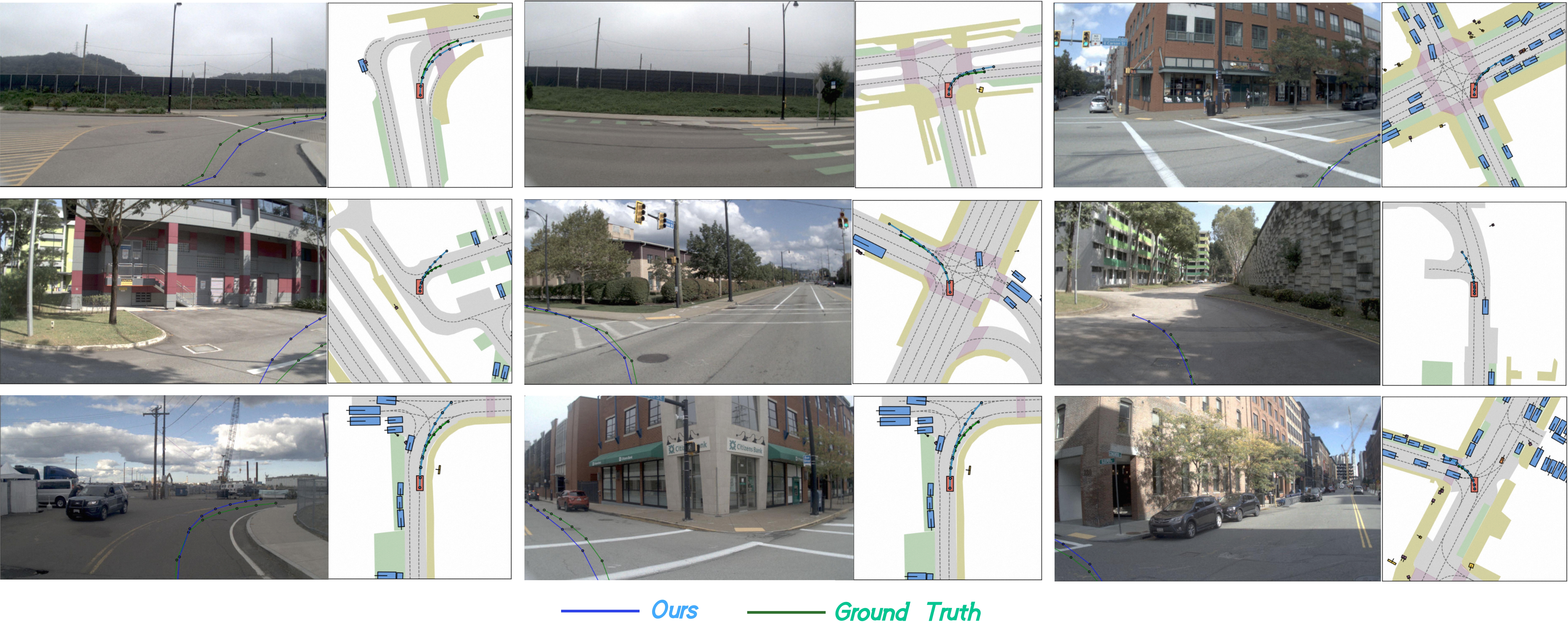}
    \caption{Representative failure cases on the NAVSIM benchmark.}
    \label{fig:bad_case}
\end{figure}

\begin{figure}[!t]
    \centering
    \includegraphics[width=1\linewidth]{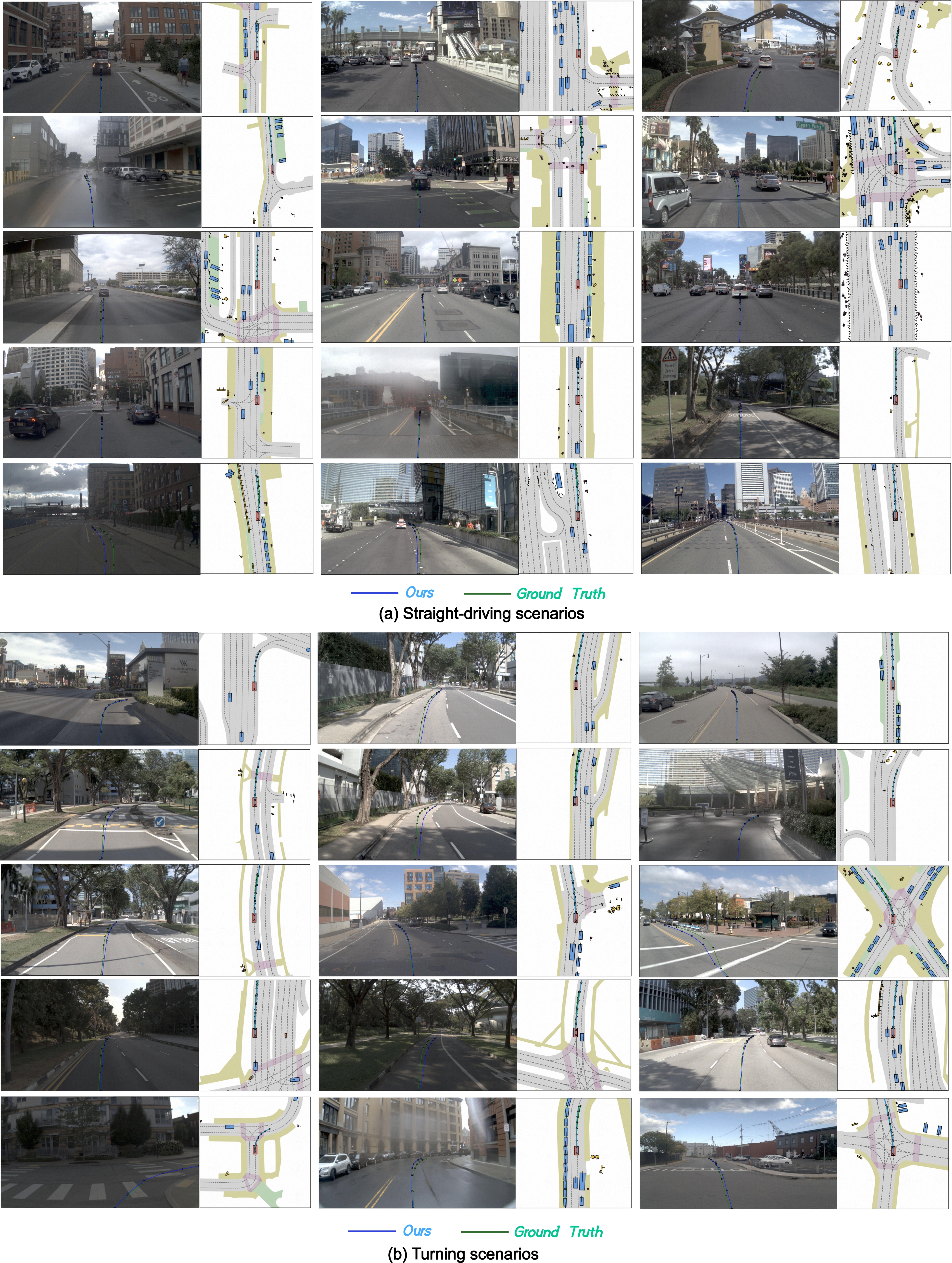}
    \caption{Additional qualitative results on the NAVSIM benchmark.}
    \label{fig:good_all}
\end{figure}

For the SFT route calibration weight $\lambda$, increasing $\lambda$ makes the initial routing distribution more balanced, thereby preserving a stronger dual-route prior. However, larger values do not further improve downstream planning. The best PDMS is obtained at $\lambda = 0.10$, suggesting that moderate route calibration is sufficient to prevent early route collapse without over-regularizing the supervised objective.

For the RL route loss weight $\lambda_r$, introducing route-level supervision clearly improves routing quality over the $\lambda_r = 0$ baseline. As $\lambda_r$ increases from 0 to 0.05, RouteAcc and AUC improve while AvgRegret decreases, and PDMS correspondingly rises from 90.9 to 91.2. At the same time, the average response length is reduced, indicating that better route learning also avoids unnecessary reasoning. When $\lambda_r$ becomes too large, both routing quality and planning performance degrade, suggesting that excessive emphasis on route-level optimization can interfere with content generation.

\rev{The thresholds $\tau_{\mathrm{low}}$, $\tau_{\mathrm{high}}$, and $\epsilon$ control allocation of the fixed 16-rollout budget. Table~\ref{tab:rl_ablation} tests this mechanism on/off, and Table~\ref{tab:rollout_dynamics} reports its training dynamics.}

\clearpage
\section{Additional qualitative results}
\label{app:qualitative}

Figure~\ref{fig:qual_compare_app} further compares STEER with ReCogDrive~\cite{li2025recogdrive}. In both examples, our method produces trajectories that better align with the road geometry and intended maneuver, whereas the baseline tends to under-steer and deviate from the desirable path. These visualizations are consistent with the quantitative gains and illustrate that STEER improves both route selection and trajectory execution.

\section{Additional visualizations}
\label{appendix:visualization}

We provide additional qualitative results in Figure~\ref{fig:good_all}. In both straight-driving and turning scenarios, the predicted trajectories closely follow the ground-truth trajectories, illustrating the effectiveness of STEER across diverse driving conditions.

We also show representative failure cases in Figure~\ref{fig:bad_case}. When the model relies on a single front-view image, prediction errors may still occur in sharp-turn scenarios with limited contextual coverage. In such cases, the restricted field of view makes accurate long-horizon planning more difficult and can occasionally lead to lane-selection errors. These cases suggest that incorporating multi-view observations may further improve robustness in future work.

\section{Limitations}
\label{appendix:limitation}

STEER has several limitations. First, our current experiments focus primarily on a single front-view input setting, which can limit route selection and planning accuracy in heavily occluded or interaction-rich scenarios. Second, the current formulation treats adaptive reasoning as a binary routing problem between a reasoning route and a direct-response route, and does not yet model finer-grained reasoning strategies such as variable reasoning depth. Third, although STEER avoids manually annotated scene-difficulty labels, route learning still depends on simulator-based reward quality. Extending the method to richer sensory inputs, more expressive routing spaces, and broader reward formulations is an important direction for future work. \rev{Evaluation is restricted to NAVSIM. Training-seed variance, entropy-threshold sensitivity, and rare high-regret routing errors remain to be studied.}

%%%%%%%%%%%%%%%%%%%%%%%%%%%%%%%%%%%%%%%%%%%%%%%%%%%%%%%%%%%%

\clearpage

\end{document}